\pdfoutput=1

\documentclass[11pt]{article}

\usepackage[preprint]{acl}

\usepackage{times}
\usepackage{latexsym}

\usepackage[T1]{fontenc}

\usepackage[utf8]{inputenc}

\usepackage{microtype}

\usepackage{inconsolata}

\usepackage{graphicx}
\usepackage{url}
\usepackage{adjustbox}
\usepackage{booktabs}
\usepackage{multirow}
\usepackage{subcaption}
\usepackage{xcolor}
\usepackage{amsfonts}
\usepackage{amsmath}
\usepackage{dsfont}
\usepackage{cleveref}

\definecolor{myBlue}{rgb}{0.0, 0.0, 1.0}
\definecolor{myGreen}{rgb}{0.0, 0.5, 0.0}

\title{Right Answer, Wrong Score: Uncovering the Inconsistencies of LLM Evaluation in Multiple-Choice Question Answering}

\author{Francesco Maria Molfese\thanks{\ Equal contribution.} \\
	Sapienza University of Rome \\
	\texttt{molfese@diag.uniroma1.it} \\\And{}
	Luca Moroni\footnotemark[1] \\
	Sapienza University of Rome \\
	\texttt{moroni@diag.uniroma1.it} \\\And{}
	Luca Gioffré \\
	Sapienza University of Rome \\
	\texttt{gioffre@diag.uniroma1.it} \\ \AND{}
	Alessandro Scirè \\
	Babelscape \\
	\texttt{scire@babelscape.com} \\\And{}
	Simone Conia \\
	Sapienza University of Rome \\
	\texttt{conia@diag.uniroma1.it} \\\And{}
	Roberto Navigli \\
	  Sapienza University of Rome \\
	\texttt{navigli@diag.uniroma1.it} \\
}

\begin{document}
\maketitle
\begin{abstract}
One of the most widely used tasks for evaluating Large Language Models (LLMs) is Multiple-Choice Question Answering (MCQA).
While open-ended question answering tasks are more challenging to evaluate, MCQA tasks are, in principle, easier to assess, as the model's answer is thought to be simple to extract and is compared directly to a set of predefined choices.
However, recent studies have started to question the reliability of MCQA evaluation, showing that multiple factors can significantly impact the reported performance of LLMs, especially when the model generates free-form text before selecting one of the answer choices.
In this work, we shed light on the inconsistencies of MCQA evaluation strategies, which can lead to inaccurate and misleading model comparisons.
We systematically analyze whether existing answer extraction methods are aligned with human judgment, and how they are influenced by answer constraints in the prompt across different domains.
Our experiments demonstrate that traditional evaluation strategies often underestimate LLM capabilities, while LLM-based answer extractors are prone to systematic errors.
Moreover, we reveal a fundamental trade-off between including format constraints in the prompt to simplify answer extraction and allowing models to generate free-form text to improve reasoning.
Our findings call for standardized evaluation methodologies and highlight the need for more reliable and consistent MCQA evaluation practices.
\end{abstract}

\section{Introduction}\label{sec:intro}
MCQA is one of the most common tasks used to evaluate LLMs across various domains, including commonsense reasoning~\cite{talmor-etal-2019-commonsenseqa, mihaylov-etal-2018-suit, bisk2019piqareasoningphysicalcommonsense, sap-etal-2019-social}, grade-school science~\cite{clark2018thinksolvedquestionanswering}, and multi-domain challenges~\cite{hendrycks2021measuringmassivemultitasklanguage, wang2024mmluprorobustchallengingmultitask, gema2025mmlu}, among others.
MCQA is straightforward: given a question and a set of answer choices, the model has to select the correct answer.
Despite the apparent simplicity of the task, the evaluation of LLMs on MCQA benchmarks is not trivial, as the model's answer has either to be extracted from its generated output, or selected based on the probabilities assigned to each answer choice.

\begin{figure}[t]
    \centering
    \includegraphics[clip, trim=40 0 40 0, width=\columnwidth]{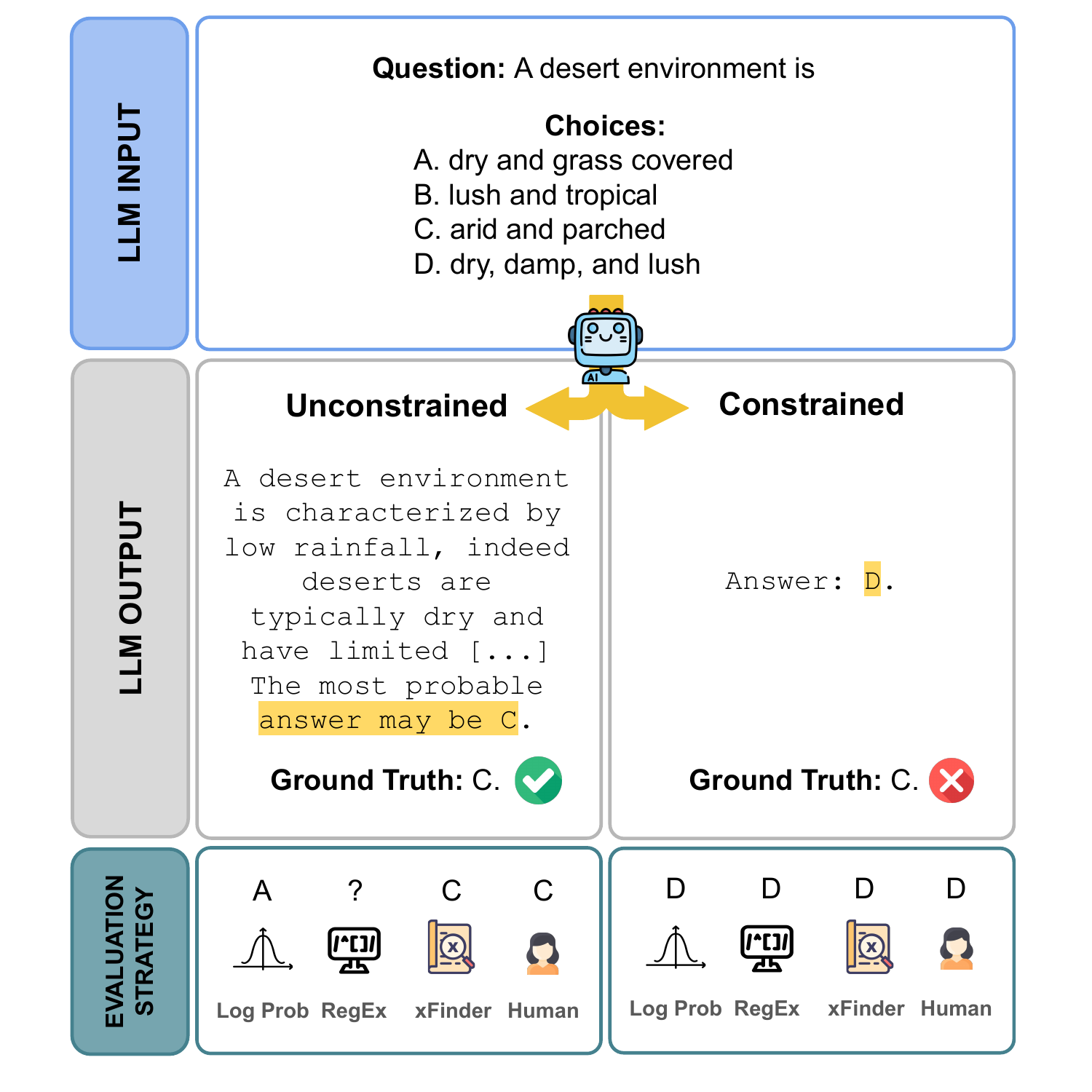}
    \caption{Different evaluation strategies (RegEx, Logprobs, xFinder and Human) and prompt settings (constrained or not) can lead to discrepancies in model performance. 
    }
    \label{img:intro}
\end{figure}

Since the introduction of techniques that enhance the reasoning capabilities of LLMs, such as Chain-of-Thought~\cite[CoT]{wei-etal-2022-cot,kojima2023largelanguagemodelszeroshot}, most models are now prompted to generate free-form text before selecting an answer, which improves their accuracy but complicates the extraction of the model's intended answer, as shown in \Cref{img:intro}.
Recently, the reliability of MCQA evaluation strategies has been called into question, as different methods can lead to significant variations in the model performance reported~\cite{wang-etal-2024-answer-c, yu2024xfinderrobustpinpointanswer}.
For example, measuring the probability of the first token generated by the model to be the label (``A'' to ``D'') of the correct answer can yield different results compared to extracting the answer from the model's output text~\cite{hendrycks2021measuringmassivemultitasklanguage, robinson2023leveraginglargelanguagemodels, zheng2024largelanguagemodelsrobust}.
As researchers introduce more sophisticated reasoning capabilities---such as test-time scaling and ``thinking'' mechanisms~\cite{deepseek-2025}---the reliability of MCQA evaluation strategies becomes increasingly important for ensuring fair comparisons.

In this work, we investigate the reliability of MCQA evaluation strategies, focusing on how different methods for extracting the model's answer impact the reported performance of LLMs.
To the best of our knowledge, we introduce the first comprehensive analysis on how different factors in the evaluation strategy, prompt setting, and benchmark domain influence model performance.
We conduct a human evaluation to assess the alignment between evaluation strategies and human judgment, highlighting the limitations and inconsistencies of current practices.
Finally, we identify cases in which existing evaluation methods systematically fail, including those methods based on LLMs, thereby highlighting which challenges still remain unsolved in MCQA evaluation.
In summary, we address the following critical research questions:
\begin{itemize}
    \item \textbf{RQ1}: How well do current evaluation strategies align with human judgment?
    \item \textbf{RQ2}: How does the choice of evaluation strategy and prompt setting impact LLM performance?
    \item \textbf{RQ3}: How does model performance shift across different benchmark domains for each prompt setting and evaluation strategy?
    \item \textbf{RQ4}: How reliable are LLM-based methods in extracting a model's intended answer?
\end{itemize}
We hope our work will lead to more rigorous and standardized evaluation practices.
We release our code and data at \url{https://github.com/SapienzaNLP/mcqa-eval}.

\section{Related Work}\label{sec:related}
This section surveys previous research in MCQA evaluation on two main aspects: how task format and inherent biases affect model performance (\Cref{sec:bias}), and how different strategies for answer extraction influence evaluation (\Cref{sec:ans-select}).

\subsection{Task Format and Label Bias}\label{sec:bias}
The research community has demonstrated that even minor variations in evaluation setup can significantly impact LLM performance.
For instance, a seminal work by \citet{robinson2023leveraginglargelanguagemodels} studied the impact of task format on LLM performance, showing that models struggle with multiple-choice symbol binding, i.e., maintaining order invariance when reasoning over different answer choices.
Building on these findings, \citet{zheng2024largelanguagemodelsrobust} documented systematic position biases in LLMs, showing that models disproportionately favor certain answer positions (e.g., ``Option A'' over ``Option B''). 
This positional sensitivity was further validated by \citet{alzahrani-etal-2024-benchmarks}, who demonstrated that reordering answer choices or modifying answer selection methods can alter leaderboard rankings.

A parallel line of research has questioned whether LLMs truly require question context for MCQA.
\citet{balepur-etal-2024-artifacts} showed that LLMs can achieve high performance on MCQA benchmarks without access to the question, suggesting that they may rely on spurious correlations in the answer choices.
Further evidence of such shortcuts emerged in \citet{wang-etal-2025-llms-may}'s work, suggesting that LLMs can select answers by eliminating clearly incorrect options rather than identifying the most accurate choice.

Unlike previous studies focused on answer selection biases and task formulation, our work examines how variations in prompt settings and format---including the trade-off between prompts that impose constraints on the answer format and those that allow free-form text generation---affect model performance across evaluation strategies.

\subsection{Evaluation Strategies}\label{sec:ans-select}
Current MCQA evaluation approaches broadly fall into two categories: those based on direct probability analysis and those that require answer extraction from the model's output.
If, given a question, there is a finite set of answer choices, we can assign a simple label to each choice (e.g., ``A'' to ``D'') and compute the next-token probability distribution for each label after ``\verb|Answer:|''.
This method, which we refer to as Logprobs, has been widely used in recent studies~\cite{hendrycks2021measuringmassivemultitasklanguage, robinson2023leveraginglargelanguagemodels, zheng2024largelanguagemodelsrobust}.
Although Logprobs is computationally efficient and conceptually straightforward, it cannot be applied when the model generates free-form text before selecting an answer, which is becoming an increasingly common practice in MCQA evaluation, e.g., with Chain-of-Thought~\cite{wei-etal-2022-cot,kojima2023largelanguagemodelszeroshot}.
In such cases, answer extraction methods are required, such as those based on regular expressions (RegEx) or LLM-based models, such as xFinder~\cite{yu2024xfinderrobustpinpointanswer}.
RegEx methods can be used to scan the model's output for predefined patterns, such as ``\verb|Answer: {label}|'' or ``\verb|The answer is: {label}|''~\cite{wang2024mmluprorobustchallengingmultitask}.
However, the effectiveness of RegEx methods requires careful crafting of patterns, which can still lead to high miss rates~\cite{yu2024xfinderrobustpinpointanswer}, especially when the model generates complex reasoning chains.
In contrast, classifier- or LLM-based methods, e.g., xFinder, are fine-tuned to extract the model's intended answer from its output, given the question and answer choices~\cite{yu2024xfinderrobustpinpointanswer}.

Our work builds on these studies by systematically analyzing the reliability of MCQA evaluation, examining how evaluation strategies, prompt settings, and benchmark domains affect performance assessment. Importantly, we evaluate the agreement between automated evaluation strategies and human judgment, providing insights into the challenges and limitations of current practices.

\section{Methodology}\label{sec:method}
Our investigation into MCQA evaluation reliability covers three dimensions: evaluation strategies, prompt settings, and benchmark domains.
For each dimension, we design controlled experiments that isolate specific variables while maintaining others constant, allowing us to measure:
\begin{itemize}
    \item The correlation between automated evaluation strategies and human judgment;
    \item The impact of prompt constraints on model reasoning and answer extraction;
    \item The variation in evaluation reliability across different domains.
\end{itemize}
In the following sections, we first formalize the MCQA task and our evaluation framework (\Cref{sec:task}), then outline the evaluation strategies under investigation (\Cref{sec:strategies}), and finally describe our prompt settings (\Cref{sec:prompts}).

\subsection{Task Formulation}\label{sec:task}
Let \(\mathcal{D} = \{(q_i, C_{q_i}, a_i)\}_{i=1}^{N}\) be a dataset of \(N\) multiple-choice instances, where each instance consists of a question \(q_i \in Q\), a set of \(k\) answer choices \(C_{q_i} = \{c_{1}, c_{2}, \dots, c_{k}\}\), and a ground-truth answer \(a_i \in C_{q_i}\).
The MCQA task requires a model \(f\) to generate an output \(t_i\) in response to a question \(q_i\) and its corresponding choices \(C_{q_i}\). Formally, we write:
\[
t_i = f\bigl(q_i, C_{q_i}\bigr),
\]
where \(t_i\) is free-form text or a structured response (e.g., ``Answer: C''). To extract the model's intended answer \(\hat{c}_i\) from its output \(t_i\), we apply an evaluation strategy \(s\), yielding:
\[
\hat{c}_i = s\bigl(t_i\bigr).
\]
The prediction \(\hat{c}_i\) is considered correct if \(\hat{c}_i = a_i\). We assess the model's overall performance on a dataset \(\mathcal{D}\) by computing the accuracy:
\[
\textrm{Acc}(f, s) \;=\; \frac{1}{N} \sum_{i=1}^{N} \mathds{1}\bigl[\hat{c}_i = a_i\bigr],
\]
where \(\mathds{1}[\cdot]\) is the indicator function that evaluates to \(1\) if its argument is true and \(0\) otherwise. 
This framework allows us to compare how different evaluation strategies \(s\) and prompting configurations influence the final accuracy of model \(f\).

\subsection{Evaluation Strategies}\label{sec:strategies}
To assess LLM performance on MCQA tasks, we need a method to identify or extract the model's intended answer $\hat{c}$ from its output $t$, given a question $q$ and its answer choices $C_q$.
We experiment with three evaluation strategies, which are representatives of traditional approaches or emerging trends in MCQA evaluation.

\paragraph{Logprobs:} Rather than extracting answers from generated text, this strategy analyzes the model's probability distribution over first tokens after a prompt terminating with $t_0 =$ ``\verb|Answer:|''~\cite{hendrycks2021measuringmassivemultitasklanguage}. Formally:
\[
    s_{\text{logprob}}(q, C_q) = \operatorname{argmax}_{c \in C_q} P(c | q, C_q, t_0)
\]
where $P(c|q,C_q,t_0)$ is obtained by applying the softmax operation to the model's log-probabilities corresponding to the labels (e.g., ``A'' to ``D'') of the choices. Although efficient, this method cannot handle free-form text generation or chain-of-thought reasoning.

\paragraph{RegEx:} This parameterless method applies a set of regular expressions $\mathcal{R} = \{r_1, \dots, r_m\}$ to extract the model's answer. For an output $t$, we define:
\begin{equation*}
    \begin{split}
        s_{\text{regex}}(t) = \begin{cases}
            \text{match}(r_i, t) & \text{if } \exists r_i \in \mathcal{R} \text{ matching } t \\
            \emptyset & \text{otherwise}
        \end{cases}
    \end{split}
\end{equation*}
where $\textrm{match}(r_i, t)$ returns the first answer choice label that matches pattern $r_i$ in $t$. Although computationally efficient, this approach can fail when models generate complex reasoning chains or deviate from expected patterns.

\paragraph{LLM-based answer extraction:} This approach uses an LLM $s_{\textrm{llm}}$ fine-tuned to extract answers from arbitrary outputs:
\[
s_{\textrm{llm}}(q, C_q, t) = \operatorname{LLM}(q, C_q, t)
\]
LLM-based approaches are relatively new and have shown promising results.
Unlike Logprobs and RegEx, LLM-based methods can handle free-form text generation and complex reasoning chains, making them more robust to variations in model output.
We evaluate two state-of-the-art models: xFinder-Llama (8B parameters) and xFinder-Qwen (500M parameters), introduced by \citet{yu2024xfinderrobustpinpointanswer}.

\vspace{\baselineskip}

Full details about the RegEx patterns used in this work, as well as the xFinder models can be found in \Cref{sec:appendix-strategies}.

\subsection{Prompt Settings}\label{sec:prompts}
We investigate how four widely-used prompt settings influence both model performance and evaluation reliability. For each setting $p \in \mathcal{P}$, we define a prompt template:
\[
\mathcal{P}_p(q, C_q) = \text{sys}_p \oplus \text{inst}_p(q, C_q) \oplus \text{const}_p
\]
where $\text{sys}_p$ is the system prompt, $\text{inst}_p$ the instruction template, and $\text{const}_p$ any format constraints. We focus on the following four settings and hypothesize that results on other settings would follow similar trends:

\paragraph{Zero-Shot (ZS):} The model receives only a system prompt, followed by the question and available choices. This setting imposes no constraints on the output format, allowing the model complete freedom in response generation.

\paragraph{Zero-Shot Chain-of-Thought (ZS-CoT):} This setting prompts the model to use CoT, allowing it to explain its reasoning before selecting the answer~\cite{kojima2023largelanguagemodelszeroshot}.

\paragraph{Zero-Shot with Format Constraint (ZS-Const):} Similar to ZS, but with a format constraint on the answer. The LLM is prompted to respond in a specific format, e.g., ``\verb|Answer: {label}|''~\cite{wang-etal-2024-answer-c}, which simplifies answer extraction.

\paragraph{Few-Shot (FS):} The model is provided with $n$ examples randomly selected from the training or validation set of the benchmark. These examples are structured as multi-turn conversations, following common practice~\cite{eval-harness}. There are no constraints on the answer, which the LLM can observe from the provided examples.

\vspace{\baselineskip}

These prompt settings allow us to study the trade-off between format constraints vs.\ free-form text generation, and LLM performance vs.\ simplicity of answer extraction. Full details of the prompts used in this work can be found in \Cref{sec:appendix-prompts}.

\section{Experimental Setup}\label{sec:exp}

\subsection{Benchmark Selection}\label{sec:datasets}
We select 3 popular MCQA benchmarks, each targeting different aspects of language understanding:

\paragraph{MMLU-Redux:} A manually curated subset of MMLU \cite{hendrycks2021measuringmassivemultitasklanguage} comprising 5,700 questions across 57 domains. This dataset addresses potential quality issues in the original MMLU by incorporating additional information and annotation provided by experts in order to review and correct problematic instances~\cite{gema2025mmlu}. The domains span four major categories: \textsc{Stem}, \textsc{Humanities}, \textsc{Social Sciences}, and \textsc{Other}.

\paragraph{OpenBookQA:} A question-answering dataset \cite[OBQA]{mihaylov-etal-2018-suit} that tests factual recall and multi-hop reasoning. Each question requires combining scientific facts with common sense.

\paragraph{ARC-Challenge:} A collection of grade-school science questions \cite[ARC]{clark2018thinksolvedquestionanswering} selected to be challenging for NLP systems. Questions often require complex reasoning and external knowledge.

\vspace{\baselineskip}

Following standard practice, we evaluate on the provided test sets. For few-shot experiments, we randomly sample five examples from training sets when available, or validation otherwise.

\subsection{Model Selection}\label{sec:models}
We evaluate eight LLMs with different architectures and sizes, ranging from 1 billion to 8 billion parameters. The models are selected to represent a diverse set of LLMs, including both high-performing and smaller, efficient models. More specifically, we evaluate the following models:
\begin{itemize}
    \item \textbf{Small-scale LLMs (1B -- 4B):} Llama-3.2-1B-Instruct, Phi-3.5-mini-instruct, Phi-4-mini-instruct and SmolLM2-1.7B-Instruct.
    \item \textbf{Medium-scale LLMs (4B -- 8B):} Llama-2-7B-chat-hf, Qwen2.5-7B-Instruct, Llama-3.1-8B-Instruct and Mistral-7B-Instruct-v0.3.
\end{itemize}
We perform greedy decoding with all models, setting the temperature to $0.0$ and allowing the models to generate a maximum of $512$ tokens.
Due to budget constraints, we exclude models with more than 8 billion parameters, hypothesizing that larger models would exhibit similar trends to those analyzed in this work.


\section{Results}\label{sec:results}

\begin{table}[t]
    \centering
    \begin{adjustbox}{max width=\columnwidth}
    \centering
        \begin{tabular}{lcccc|c}
            \toprule
            & \multicolumn{5}{c}{\textbf{Agreement with Humans}} \\
            \cmidrule{2-6}
            \textbf{Eval. Strategy} & \textbf{ZS} & \textbf{ZS-CoT} & \textbf{ZS-Const} & \textbf{FS} & \textbf{Avg.} \\
            \midrule
            \textbf{RegEx} & 90.7 & 84.3 & 97.9 & 97.3 & 92.5 \\
            \textbf{Logprobs} & 74.7 & --- & 94.1 & 90.4 & ~~86.4* \\
            \textbf{xFinder-Llama} & 95.8 & 89.7 & 98.4 & 97.3 & 95.3 \\
            \textbf{xFinder-Qwen} & 94.8 & 90.3 & 98.4 & 97.3 & 95.2  \\
            \midrule
            \textbf{Human} & 98.2 & 97.0 & 98.7 & 100.0~~ & 98.5\\
            \bottomrule
        \end{tabular}
    \end{adjustbox}
    \caption{Average agreement between human annotators and evaluation strategies across eight LLMs in each prompt setting, measured with Cohen's kappa. For Logprobs, the average agreement (marked with *) is computed over ZS, ZS-Const and FS, excluding ZS-CoT.}
    \label{tab:agreement}
\end{table}

\paragraph{RQ1: ``\textit{How well do current evaluation strategies align with human judgment?}''}
To answer this question, we conduct a manual annotation process in which human annotators extract the intended answer from the model's response across all prompt settings and evaluation strategies. 
First, we prompt each of the eight LLMs on the full MMLU-Redux dataset under four different prompt settings. We then randomly sample a total of 1,000 \( (q, C_q) \) instances for annotation, ensuring a balanced distribution across the different prompt settings.
Then, four human annotators extract the intended answer from the model's response, assigning a label from ``A'' to ``D'' or a special tag, ``[No valid answer],'' for cases where the model produces an invalid response. If a response is invalid, annotators are required to specify the reason for invalidity, which can arise from various factors, including: i) conflicting answers (e.g., the reasoning produced by the model supports choice ``C'' but the model concludes with ``Answer: A''), ii) label binding inconsistencies (e.g., the model responds with ``Answer: C. bank'' where ``bank'' corresponds to option ``B''), iii) refusal to answer (e.g., due to safety concerns or insufficient knowledge), iv) irrelevant response, where the model fails to engage with the question, and v) generation limits, e.g., the model generates a response that exceeds the token limit. In total, each annotator is assigned $400$ instances, with $200$ instances shared between all annotators in order to assess inter-annotator agreement.
We provide details on the annotation process in Appendix \ref{sec:appendix-ann}.

We compute the agreement between annotators using pairwise Cohen's kappa, averaging across all pairs, yielding a score of $98.5$, indicating an ``almost perfect'' agreement. This shows that human annotators are consistent in extracting the intended answer from the model's response, providing a reliable benchmark for evaluating the alignment between automated evaluation and human judgment.

Having created a gold dataset of $1,000$ instances, we evaluate the agreement between human annotators and automated evaluation strategies using Cohen's kappa\footnote{We used majority voting for the $200$ shared instances and the single available annotation for the remaining $800$.}. The results are reported in \Cref{tab:agreement}. We observe that LLM-based approaches for answer extraction generally achieve higher agreement with human judgment compared to traditional methods. In particular, xFinder-Llama displays the highest agreement with humans across all prompt settings, outperforming traditional strategies, namely RegEx and Logprobs, by a significant margin. However, the agreement between humans and evaluation strategies is not consistent across prompt settings. Recent work on LLMs is moving away from constrained prompts so as to allow models to generate free-form text with a view to improving reasoning. But our analysis shows that moving from ZS-Const to ZS leads to a significant drop in agreement between humans and evaluation strategies: $-2.6$\% for xFinder-Llama (from $98.4$ to $95.8$), $-3.6$\% for xFinder-Qwen (from $98.4$ to $94.8$), $-7.2$\% for RegEx (from $97.9$ to $90.7$), and $-19.2$\% for Logprobs (from $94.1$ to $74.7$). This is even more pronounced in the ZS-CoT setting, where the agreement between xFinder-Llama---the best model on average, with 8B parameters---and humans drops by $8.7$\%.
In contrast, pairwise human agreement shows just a minor drop when changing the prompt setting from ZS-Const to ZS ($-0.5$\%) or from ZS-Const to ZS-CoT ($-1.7$\%).
These results suggest that the extent to which models adhere to the required format has a significant impact on the reliability of the evaluation strategy employed, and also that state-of-the-art LLM-based approaches are not immune to this variability, especially in settings where models generate free text.

\begin{figure}[t]
    \centering
    \begin{subfigure}{\columnwidth}
        \centering
        \includegraphics[width=0.95\columnwidth]{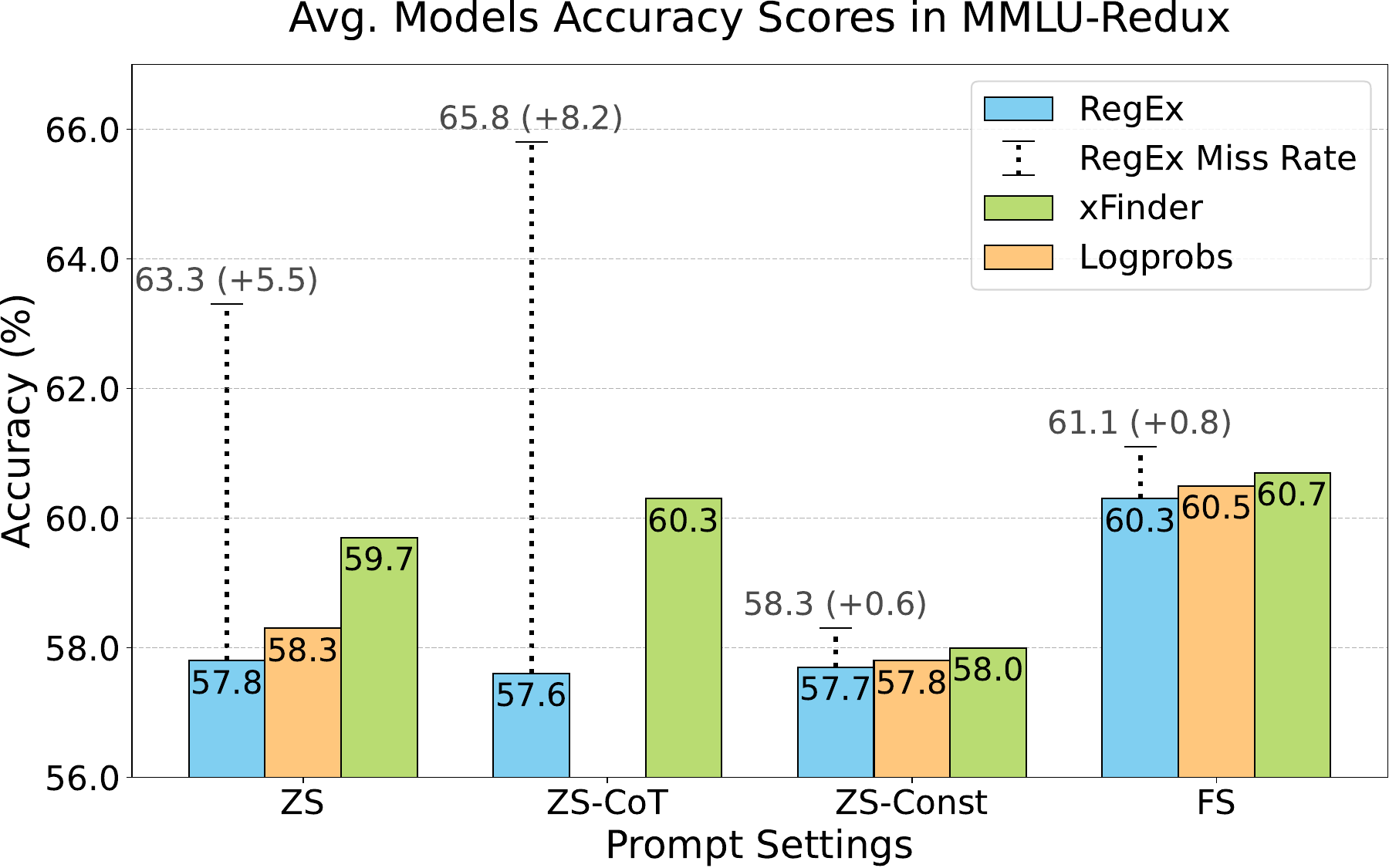}
        \caption{MMLU-Redux dataset}
        \label{img:mmlu}
    \end{subfigure}

    \vspace{0.5em}
    
    \begin{subfigure}{\columnwidth}
        \centering
        \includegraphics[width=0.95\columnwidth]{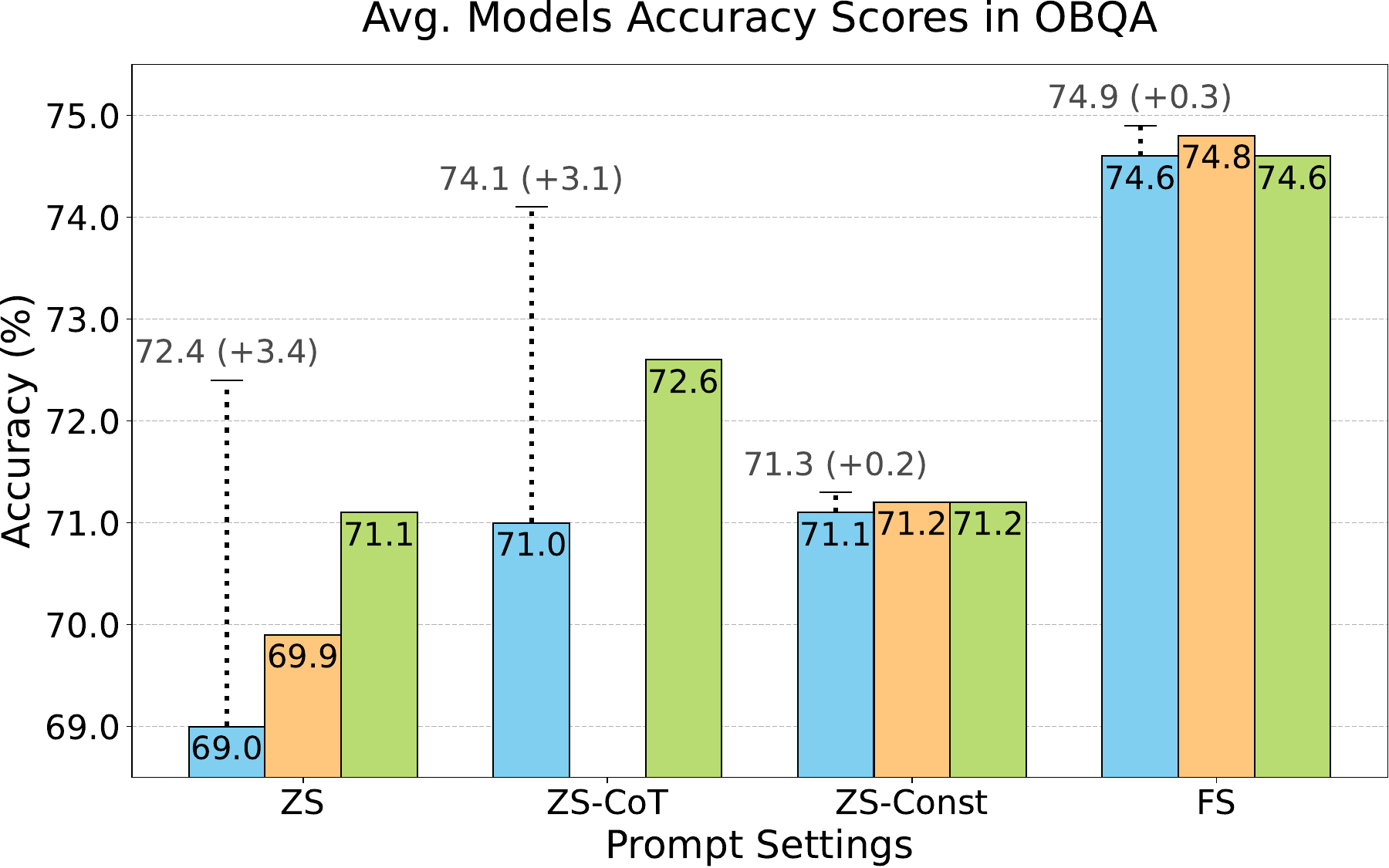}
        \caption{OBQA dataset}
        \label{img:obqa}
    \end{subfigure}
    
    \caption{Average accuracy scores across eight LLMs and four prompt settings when evaluated on the MMLU-Redux (\Cref{img:mmlu}) and OBQA (\Cref{img:obqa}) datasets. Dotted lines indicate the RegEx miss rate.}
    \label{fig:combined}
\end{figure}

\begin{figure}[t]
    \centering
    
    \begin{subfigure}{\columnwidth}
    \includegraphics[width=\columnwidth]{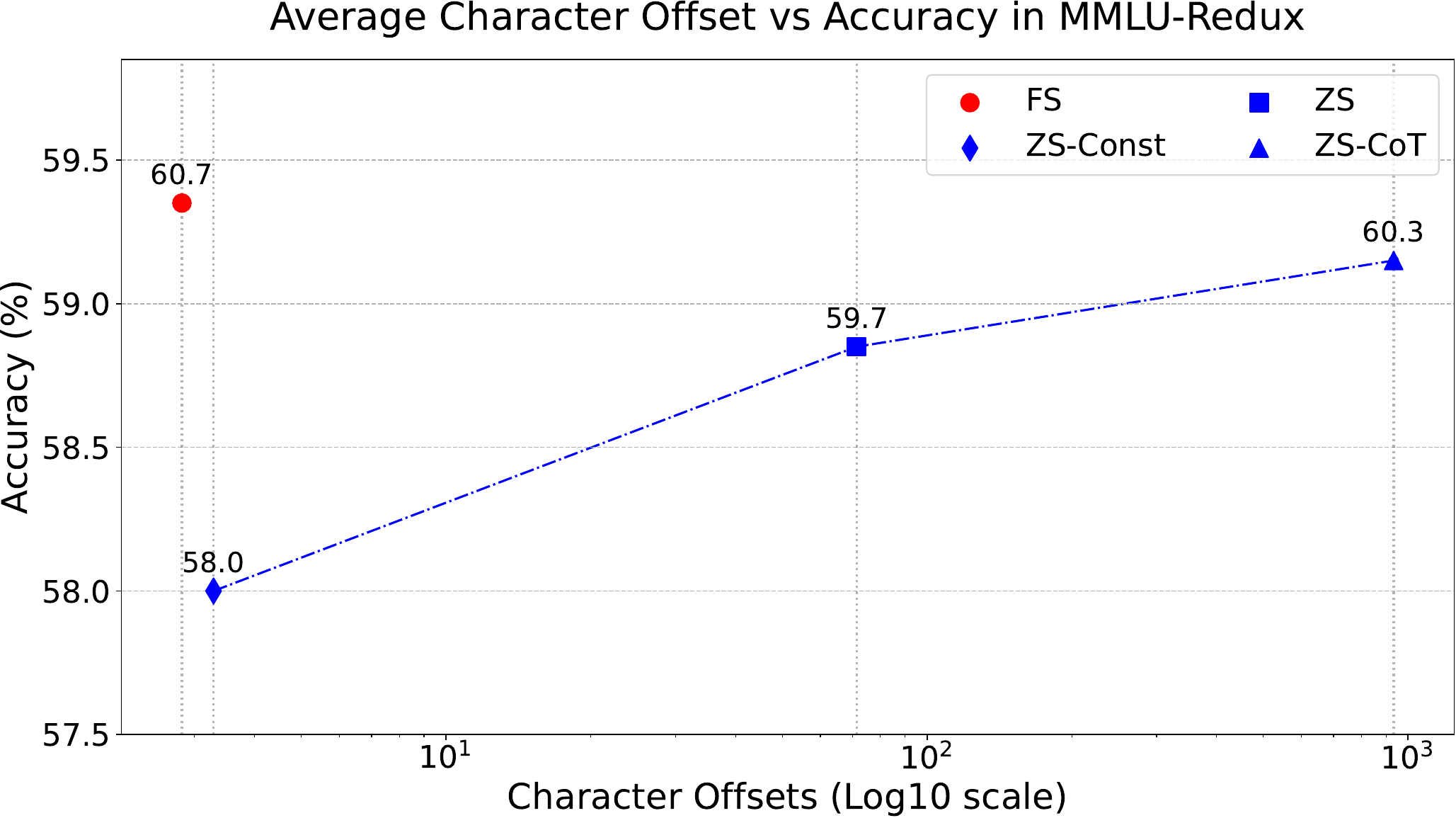}
    \caption{MMLU-Redux dataset}
    \label{fig:mmlu-offset}
    \end{subfigure}
    
    \begin{subfigure}{\columnwidth}
    \includegraphics[width=\columnwidth]{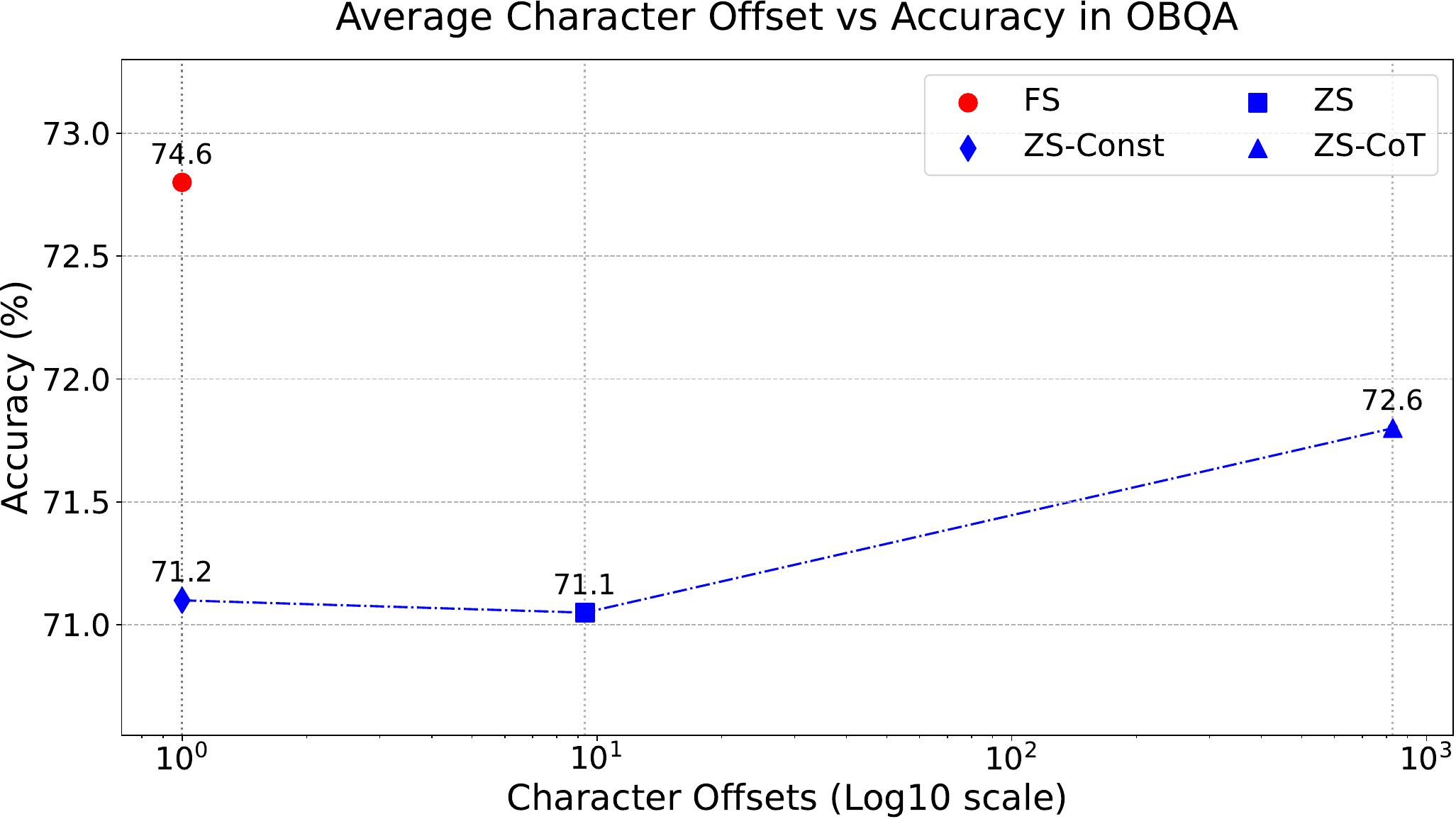}
    \caption{OBQA dataset}
    \label{fig:obqa-offset}
    \end{subfigure}
    
    \caption{The plots show the relationship between average answer offset ($\log_{10}$ scale) and accuracy (\%) for different settings using the xFinder evaluation strategy.}
    \label{img:offset-comb}
\end{figure}

\paragraph{RQ2: ``\textit{How does the choice of evaluation strategy and prompt setting impact LLM performance?}''}
The disagreement between human annotators and evaluation strategies shown in \Cref{tab:agreement} raises the question of how the choice of evaluation strategy and prompt setting affects LLM performance.
Therefore, we analyze the behavior of the eight LLMs on MMLU-Redux, OBQA, and ARC using the four prompt settings and the three evaluation strategies.
\Cref{img:mmlu,img:obqa} present the results on the MMLU-Redux and OBQA datasets, respectively. Additionally, we provide the results on the ARC dataset and the individual performance of all the LLMs in \Cref{sec:appendix-arc,sec:appendix-each-llm}.

The plots show that in prompt settings that constrain the output format---either explicitly, as in ZS-Const, or implicitly through few-shot examples, as in FS---LLMs performance remain stable across evaluation strategies, which is consistent with the high agreement between humans and evaluation strategies in these settings, i.e., simplifying answer extraction leads to more reliable evaluation outcomes.
However, our results also show that a simplified evaluation process can hide the true capabilities of current LLMs, as models in ZS or ZS-CoT generate outputs in which the answer is harder to extract, e.g., leading to a higher RegEx miss rate.\footnote{We define RegEx miss rate as the percentage of instances where no RegEx pattern is able to extract an answer from the model output.}
Interestingly, the prompt settings that show the largest differences between the results obtained with different evaluation strategies are the ones where the disagreement with human annotators is the highest, i.e., ZS and ZS-CoT.
As the research community moves towards letting models generate more complex free text before selecting an answer, current evaluation strategies are likely to become less reliable.

To assess the trade-off between generating longer responses and model performance, we study the improvement in model performance as the average answer offset increases. The average answer offset is defined as the number of characters after which one of the available RegEx patterns matches the model's intended answer.
\Cref{img:offset-comb} shows that, while higher offsets generally lead to better performance, the gain is often marginal beyond a certain threshold, e.g., moving from $10^2$ to $10^3$ characters only provides a +0.6\% improvement in accuracy in MMLU-Redux\footnote{For OBQA, the +1.2\% improvement in accuracy corresponds to only six instances.}. Given the agreement study reported in \Cref{tab:agreement}, this suggests that even LLM-based methods for answer extraction can struggle to generalize to longer responses, highlighting the need for more robust evaluation strategies.

\begin{figure}[t]
    \centering
    \begin{subfigure}{\columnwidth}
        \centering
        \includegraphics[width=0.95\columnwidth]{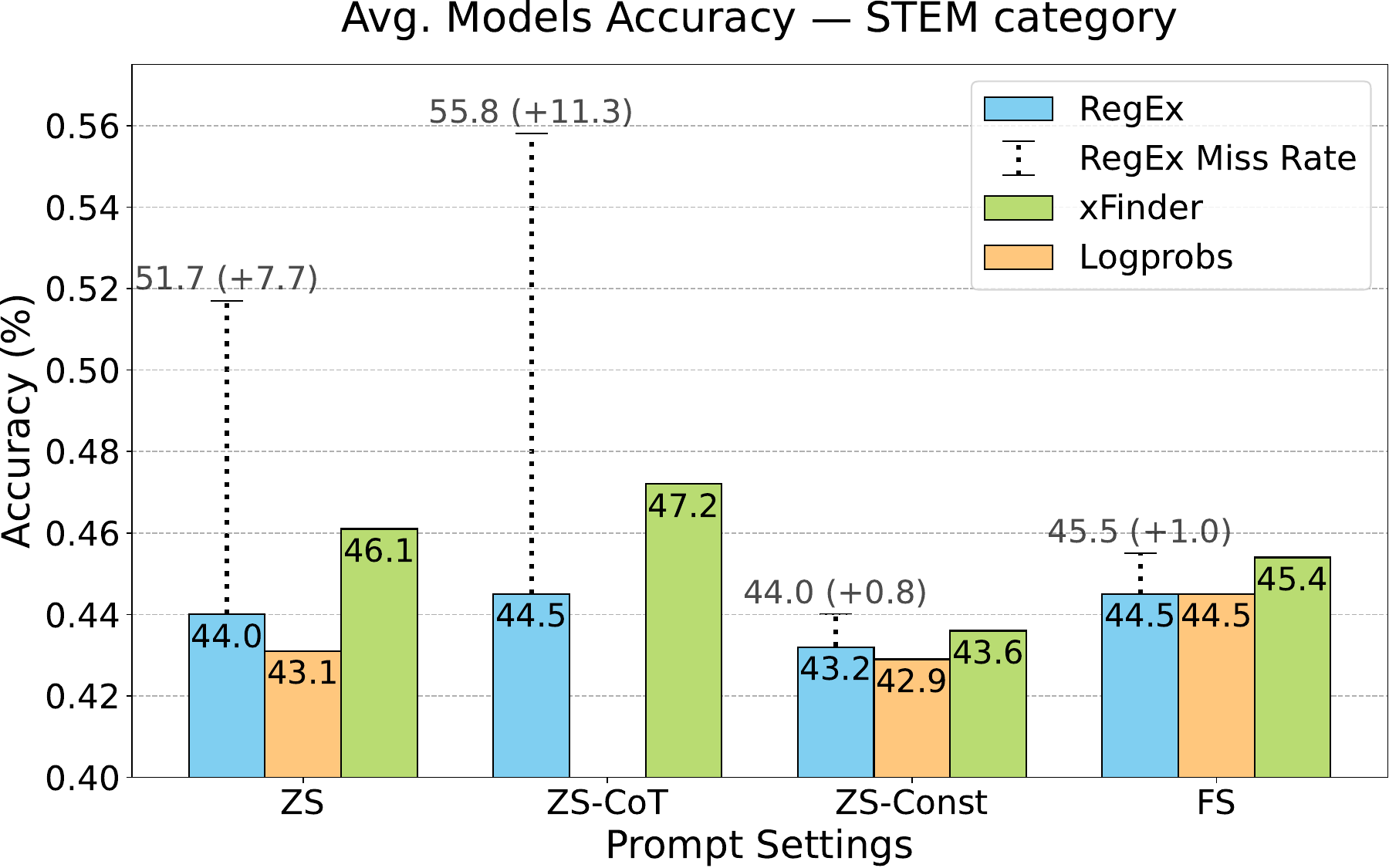}
        \caption{\textsc{Stem} category}
        \label{img:stem}
    \end{subfigure}

    \vspace{1em}
    
    \begin{subfigure}{\columnwidth}
        \centering
        \includegraphics[width=0.95\columnwidth]{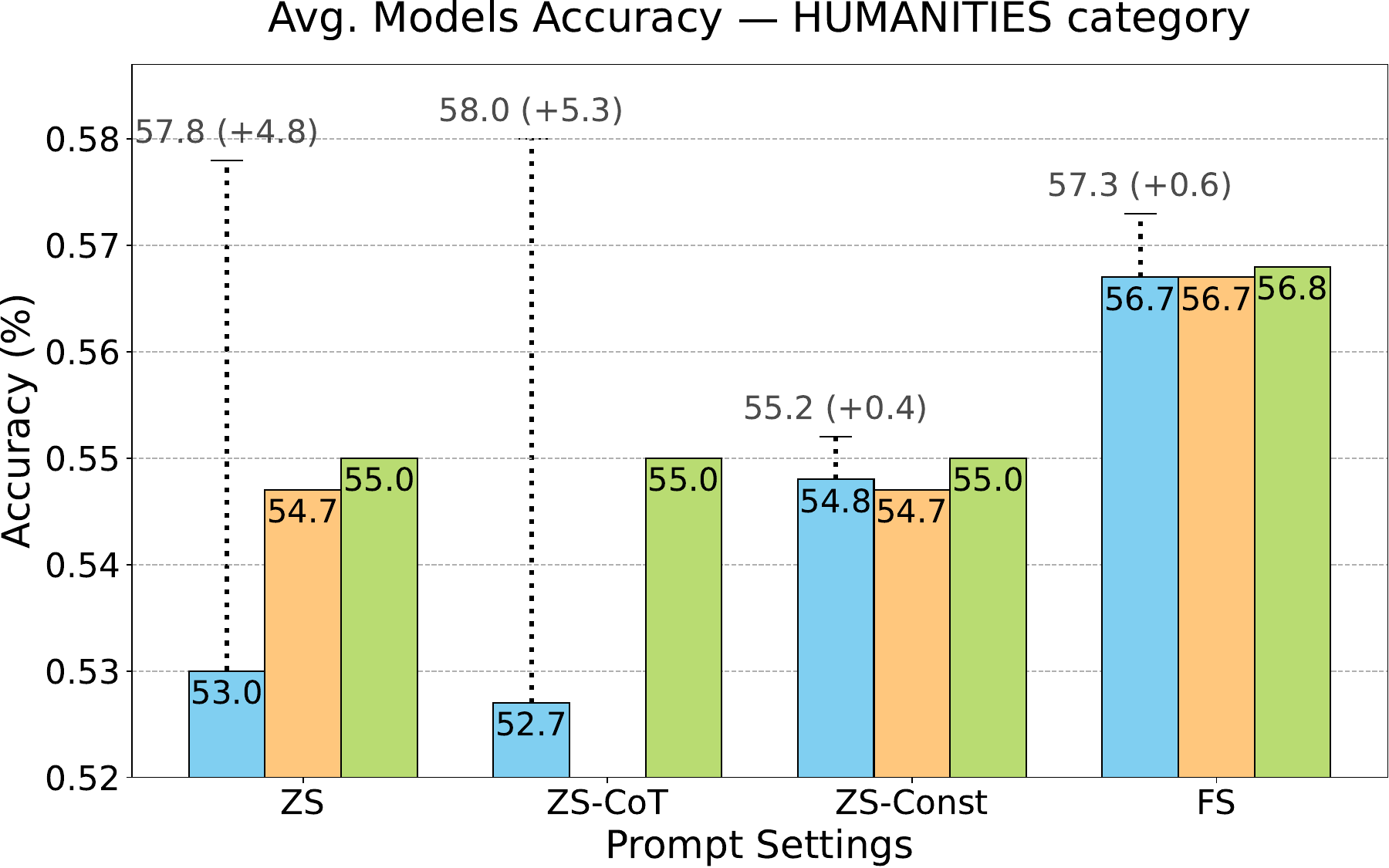}
        \caption{\textsc{Humanities} category}
        \label{img:humanities}
    \end{subfigure}
    
    \caption{Average accuracy scores across eight LLMs and four prompt settings when evaluated on the \textsc{Stem} (\Cref{img:stem}) and \textsc{Humanities} (\Cref{img:humanities}) categories of the MMLU-Redux dataset. Dotted lines indicate the RegEx miss rate.}
    \label{fig:supercategories}
\end{figure}

\definecolor{nicegreen}{RGB}{192,239,171}
\definecolor{niceblue}{RGB}{174,224,246}
\definecolor{nicered}{RGB}{239,171,192}
\fboxsep0.1em

\begin{table*}[t]
    \begin{adjustbox}{width=\textwidth}
    \scriptsize 
    \centering
    \begin{tabular*}{\textwidth}{p{0.40\textwidth}|p{0.55\textwidth}}
        \toprule
        \textbf{Question} & \textbf{Model output containing inconsistent reasoning or multiple answers} \\ 
        \midrule
        \parbox[t]{0.40\textwidth}{
        On which continent are most of the Venus figurines found?\\
        A. Europe $\quad$ B. Africa $\quad\quad$ C. Asia $\quad\quad$ D. Australia}        
        & 
        \parbox[t]{0.55\textwidth}{
        \texttt{[...] \colorbox{niceblue}{The majority of Venus figurines have been found on the continent} \colorbox{niceblue}{of Europe}.
        Therefore, the answer to the question is: \colorbox{nicered}{Answer: B. Africa}}}\\
        
        \midrule

        \parbox[t]{0.40\textwidth}{
        Which of the following capital cities is located OUTSIDE the central core area of the country?\\
        $\quad$ A. Abuja $\quad\ \ $ B. Paris
        $\quad\quad\ \ $ C. London $\quad$ D. Cairo
        } 
        & 
        \parbox[t]{0.55\textwidth}{
        \texttt{[...] analyze the options:  
        \colorbox{niceblue}{A. Abuja: located in the central core area} 
        [...] 
        \colorbox{niceblue}{D. Cairo: located in the central core area}
        [...] the capital city NOT located within the central core area of the country is: \colorbox{nicered}{D. Cairo}.
        }} \\
        
        \midrule
        \parbox[t]{0.40\textwidth}{
        Question: If $f (x) = x^3 + 2x^2 - 9x - 18$, which of the following statements is true?\\
        Choices:\\ 
        A. $f (x) = 0$ has three real solutions.\\ 
        B. $f (x) \geq -18$ for all $x \geq 0$.\\ 
        C. $f (x) \leq -18$ for all $x \leq 0$.\\
        D. The function $f (x)$ is decreasing for $x \leq -3$.}
        &
        \parbox[t]{0.55\textwidth}{
        \texttt{
        A. f(x) = 0 has three real solutions: [...] So, \colorbox{nicered}{the correct answer is A}.\\  
        B. f(x) >= -18 for all x >= 0: [...] Since 1782 is greater than -18, \colorbox{nicered}{the correct answer is B}.\\  
        C. f(x) <= -18 for all x <= 0:  [...] Since -64 is less than -18, \colorbox{nicered}{the correct answer is C}.\\  
        D. The function f(x) is decreasing for x <= -3:  We can determine [...] \colorbox{nicered}{the correct answer is D}.}}\\

        
        \bottomrule
    \end{tabular*}
    \end{adjustbox}
    \caption{Examples where xFinder fails to detect inconsistencies in model outputs. 
    Blue highlights the model's reasoning path towards a particular answer, while red highlights the final answer(s) selected by the model.}
    \label{tab:qualitative}
\end{table*}

\paragraph{RQ3: ``\textit{How does model performance shift across different benchmark domains for each prompt setting and evaluation strategy?}''}
To systematically analyze domain-specific effects, we use the existing categorization of MMLU-Redux, which divides the questions into four macro-domains: \textsc{Stem}, \textsc{Humanities}, \textsc{Social Sciences}, and \textsc{Other}.
We focus our analysis on \textsc{Stem} and \textsc{Humanities}, as our results show that these categories exhibit the most significant differences in model performance across prompt settings and evaluation strategies (results for other categories are provided in \Cref{sec:appendix-sub}).

As our results show in \Cref{img:stem}, models in the \textsc{Stem} category tend to perform best in the ZS and ZS-CoT settings.
In particular, we observe that, when evaluated with RegEx, LLMs achieve better scores in these settings compared to ZS-Const, despite a persistently higher miss rate.
The same holds for xFinder, where the performance gap between the ZS and ZS-Const settings increases to $2.5$ accuracy points—substantially larger than the $1.7$-point gap observed in \Cref{img:mmlu} across all MMLU domains.
This is especially important when considered jointly with our study on the agreement between human annotators and evaluation strategies: given the fact that LLM-based answer extraction strategies remain imperfect (as shown in \Cref{tab:agreement}, the agreement between xFinder and humans is $7.3$ points lower than the agreement between humans), the true performance of models in the \textsc{Stem} category may be even higher than what we report here, which calls for further research on more reliable evaluation strategies. 
This also underscores the importance of allowing models to generate freely before producing a final answer, particularly in scenarios requiring reasoning over mathematical expressions or scientific formulas.
On the other hand, models in the \textsc{Humanities} category tend to perform similarly across prompt settings, as shown in \Cref{img:humanities}, with the FS setting providing an advantage in terms of final accuracy, suggesting that for humanities-related questions, models do not need to generate long-form reasoning to provide accurate answers, but rather rely on recalling factual information stored in their parameters.

\paragraph{RQ4: ``\textit{How reliable are LLM-based approaches in extracting the models' intended answers?}''}
Our experiments show that LLM-based methods like xFinder correlate well with human judgment (Table~\ref{tab:agreement}), but still underperform compared to human annotator agreement, especially on longer responses.
Here we investigate the main failure modes of xFinder to identify vulnerabilities and areas for improvement in state-of-the-art evaluation strategies.
In order to do so, we manually inspect the cases where xFinder and human annotators disagree.
The main source of disagreement occurs when xFinder assigns a valid answer to a model output while human annotators label it as ``[No valid answer]''.\footnote{xFinder is also trained to recognize invalid outputs and tag them with ``[No valid answer]''.}

Specifically, we identify two main patterns that consistently mislead xFinder: i) a reasoning path that supports one answer but concludes with another without justification, and ii) situations where the model presents conflicting reasoning, implying multiple answers.
We refer to these cases as ``\textit{inconsistent reasoning}'' and ``\textit{multiple answers}'' and provide examples in Table~\ref{tab:qualitative}, where blue highlights the model's reasoning path towards a particular answer, while red highlights the final answer(s) selected by the model.

The identification of these two patterns allows us to construct an adversarial dataset derived from MMLU-Redux, which we call MMLU-Adversarial.
With this adversarial dataset we aim to assess the ability of current LLM-based techniques to identify instances where the model generates invalid answers and provide the opportunity for future work to benchmark new LLM-based answer extraction methods on more challenging instances.

For the inconsistent reasoning pattern, we start from the ZS outputs of the models on MMLU-Redux and then prompt Gemini-1.5-Flash to preserve the original reasoning and swap the final answer with one that contradicts the reasoning.
For the multiple answers pattern, we generate adversarial instances by taking the original $(q, C_q)$ pairs and asking the model to explicitly generate a series of reasoning paths that motivate, explain or justify multiple answers.
The complete prompts used for dataset creation, along with input-output examples, are provided in \Cref{sec:appendix-mmlu-adv}.

To ensure the relevance and quality of the generated resource, we conduct a thorough manual verification process. This involves carefully reviewing each instance to eliminate any artifacts or unintended errors introduced during generation. As a result, we retain a curated subset consisting of 1,000 high-quality and correctly modified instances for each of the two patterns.

When evaluated on MMLU-Adversarial, xFinder-Llama correctly identifies only 1.9\% of instances exhibiting inconsistent reasoning as ``[No valid answer]'' and just 10.9\% of instances involving multiple answers. The performance of xFinder-Qwen is even more limited, correctly labeling only 0.6\% of inconsistent reasoning cases and 0.9\% of multiple answers cases. These results support our hypothesis that current LLM-based detection methods struggle to reliably identify conflicting or ambiguous reasoning in model outputs.
We argue that to justify their computational cost over parameterless alternatives, LLM-based methods should reliably tag erroneous outputs as “[No valid answer].” Such alignment with human judgment is essential to avoid inflated performance and ensure reliable automated evaluation.

\section{Can LLM-based Answer Extractors Solve the MCQA Task?}\label{sec:adversarial}
Our analysis in \Cref{sec:results} uncovered discrepancies between xFinder’s outputs and human annotations.
Additionally, in cases where the annotators labeled responses as “[No valid answer]”, we found that, in some instances, xFinder still assigned one of the available labels.
This suggests that xFinder may inherit biases from its underlying base model, occasionally attempting to solve the MCQA task rather than strictly adhering to the intended answer extraction objective for which it was fine-tuned.

To test this hypothesis, we designed three distinct prompts to highlight this unintended behavior (\Cref{sec:appendix-adv}) .
The rationale behind these prompts is the creation of an ambiguous answer that tries to stimulate the answer extractor model to solve the MCQA task.
According to xFinder's design principles, all these prompts should result in a ``[No valid answer]'' response, since no single answer is explicitly deemed as correct.

To quantify the effects that these prompts have on the behavior of xFinder, we designed two metrics: the \textit{adversarial rate} and the \textit{relative accuracy}.
The adversarial rate is defined as the percentage of instances where xFinder assigns one of the available labels instead of generating ``[No valid answer]'', while the relative accuracy is defined as the percentage of cases where xFinder selects a label that correctly matches the ground truth for a given sample in the dataset.

The results in \Cref{tab:adversarial_xfinder_all} show that xFinder is indeed prone to solving the MCQA task rather than strictly performing answer extraction. For instance, xFinder-Qwen reaches an adversarial rate of up to $96.9\%$ on the MMLU-Redux dataset, while xFinder-Llama reaches a relative accuracy of up to $89.9\%$ on OBQA and ARC. This suggests that,  when prompted adversarially, xFinder models may shift towards solving the original MCQA task instead of extracting the intended answer.


\begin{table}[t]
    \centering
    \begin{subtable}[t]{\columnwidth}
        \centering
        \begin{adjustbox}{max width=\textwidth}
            \begin{tabular}{l|cc|cc}
                \toprule
                 & \multicolumn{2}{c}{\textbf{xFinder-Llama}} & \multicolumn{2}{c}{\textbf{xFinder-Qwen}} \\
                \cmidrule{2-5}
                \textbf{Prompt} & \textbf{Adv. Rate} & \textbf{Rel. Acc.} & \textbf{Adv. Rate} & \textbf{Rel. Acc.} \\
                \midrule
                \textbf{Prompt A} & 58.9 & 68.2 & 45.7 & 29.3 \\
                \textbf{Prompt B} & 54.0 & 69.6 & 43.2 & 28.0 \\
                \textbf{Prompt C} & 15.3 & 74.8 & 96.9 & 23.1 \\
                \bottomrule
            \end{tabular}
        \end{adjustbox}
        \caption{MMLU-Redux}
        \label{tab:mmlu-adversarial_xfinder}
    \end{subtable}
    
    \vspace{1em}
    
    \begin{subtable}[t]{\columnwidth}
        \centering
        \begin{adjustbox}{max width=\textwidth}
            \begin{tabular}{l|cc|cc}
                \toprule
                 & \multicolumn{2}{c}{\textbf{xFinder-Llama}} & \multicolumn{2}{c}{\textbf{xFinder-Qwen}} \\
                \cmidrule{2-5}
                \textbf{Prompt} & \textbf{Adv. Rate} & \textbf{Rel. Acc.} & \textbf{Adv. Rate} & \textbf{Rel. Acc.} \\
                \midrule
                \textbf{Prompt A} & 49.8 & 76.7 & 19.8 & 42.4 \\
                \textbf{Prompt B} & 36.4 & 82.4 & 16.4 & 42.7 \\
                \textbf{Prompt C} & ~~5.8 & 89.9 & 92.0 & 30.0 \\
                \bottomrule
            \end{tabular}
        \end{adjustbox}
        \caption{OBQA}
        \label{tab:obqa-adversarial_xfinder}
    \end{subtable}
    
    \vspace{1em}
    
    \begin{subtable}[t]{\columnwidth}
        \centering
        \begin{adjustbox}{max width=\textwidth}
            \begin{tabular}{l|cc|cc}
                \toprule
                 & \multicolumn{2}{c}{\textbf{xFinder-Llama}} & \multicolumn{2}{c}{\textbf{xFinder-Qwen}} \\
                \cmidrule{2-5}
                \textbf{Prompt} & \textbf{Adv. Rate} & \textbf{Rel. Acc.} & \textbf{Adv. Rate} & \textbf{Rel. Acc.} \\
                \midrule
                \textbf{Prompt A} & 67.6 & 82.3 & 42.4 & 27.2 \\
                \textbf{Prompt B} & 60.1 & 84.0 & 40.0 & 26.2 \\
                \textbf{Prompt C} & 13.5 & 89.9 & 96.8 & 23.2 \\
                \bottomrule
            \end{tabular}
        \end{adjustbox}
        \caption{ARC}
        \label{tab:arc-adversarial_xfinder}
    \end{subtable}
    \caption{Results of xFinder models with adversarial prompts on three datasets. The adversarial rate indicates the percentage of instances where xFinder assigns one of the available labels instead of ``[No valid answer].'' The relative accuracy column reflects the percentage of cases where xFinder selects a label that correctly matches the ground truth for that sample in the dataset.}
    \label{tab:adversarial_xfinder_all}
\end{table}

\section{Conclusions}\label{sec:conclusions}
In this paper, we analyzed the evaluation of Large Language Models in Multiple-Choice Question Answering, examining the impact of evaluation strategies, prompt constraints, and benchmark domains on model performance. 
Our findings show that traditional RegEx-based and first-token probability approaches often underestimate model reasoning, while LLM-based extraction methods, though more aligned with human judgment, remain prone to systematic errors.
Moreover, constrained prompts improve evaluation consistency but may hinder reasoning, whereas unconstrained settings tend to enhance a model's performance, but complicate answer extraction. 
Additionally, performance varies by domain, with STEM tasks benefiting from free-form reasoning, while accuracy on humanities-related questions remains stable.
Finally, our adversarial analysis reveals that even state-of-the-art answer extractors struggle with inconsistencies in LLM-generated reasoning, underscoring the need for better verification mechanisms. 

Our analyses -- which corroborate the lesson of \citet{tedeschi-etal-2023-whats} -- highlight the need for standardized evaluation protocols in order to mitigate biases introduced by prompt constraints and answer extraction techniques. 
We hope that our findings may help researchers establish more accurate, fair, and reliable model assessments.

\section*{Limitations}\label{sec:limitations}
There are several aspects of our work that leave room for future improvements. 
First, our study is limited to English-only benchmarks. 
Expanding to multilingual and cross-lingual settings would be valuable, especially since strategies for extracting answers using LLMs in multilingual contexts remain underexplored and current multilingual LLMs are known to be English-centric in several aspects, including their naturalness~\cite{guo2024largelanguagemodelsenglish} and the composition of their vocabulary~\cite{moroni-etal-2025-optimizing}.
Second, we focus on a specific set of models and do not include other families like Mistral or Gemma. Future work could broaden the analysis to these and other models so as to better understand differences across architectures.
Third, our evaluation covers only three MCQA benchmarks. Extending this to additional datasets or tasks, including those that require more complex reasoning, dealing with adversarial examples or knowledge-intensive processes~\cite{scire2024truthmirageendtoendfactuality}, could provide deeper insights.
Fourth, due to budget constraints, we evaluate only models with parameter sizes ranging from 1 billion to 8 billion. Further expanding our analysis to models with larger parameter counts, as well as RAG systems for MCQA~\cite{molfese-etal-2024-zebra}, would be valuable.

\section*{Acknowledgments}
\begin{center}
\noindent
    \begin{minipage}{0.1\linewidth}
        \begin{center}
            \includegraphics[scale=0.05]{fair.pdf}
        \end{center}
    \end{minipage}
    \hspace{0.01\linewidth}
    \begin{minipage}{0.70\linewidth}
        Roberto Navigli and Simone Conia gratefully acknowledge the support of the PNRR MUR project PE0000013-FAIR. Simone's fellowship is fully funded by this project.
    \end{minipage}
    \hspace{0.01\linewidth}
    \begin{minipage}{0.1\linewidth}
        \begin{center}
            \includegraphics[scale=0.08]{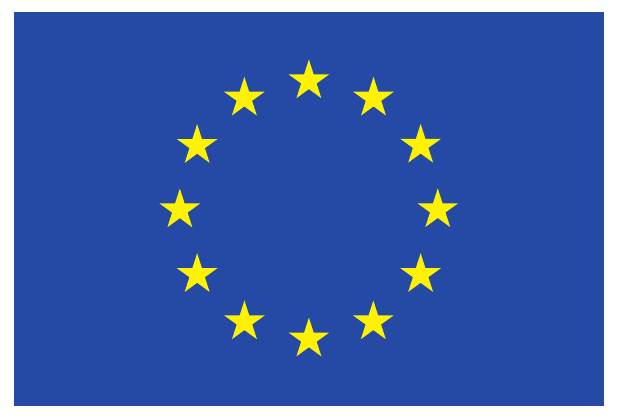}
        \end{center}
    \end{minipage}\\
\end{center}
\vspace{0.2cm}

\bibliography{ms}

\appendix

\section{Appendix}\label{sec:appendix}

\subsection{Evaluation Strategies Details}\label{sec:appendix-strategies}

In this section we present the details of our evaluation strategies, including the regular expressions we use to parse the LLMs outputs and the details of the xFinder models \cite{yu2024xfinderrobustpinpointanswer}.

\subsubsection{Regular Expressions}
To cover multiple cases, we sample several generated answers and we tune the RegEx to match the most common answer types, resulting in 18 regex patterns (\Cref{tab:regex}). 
We test each pattern sequentially on the generated output and consider only the first match for our statistics.
\begin{table*}[t]
    \scriptsize
    \begin{adjustbox}{width=\textwidth}
    \centering
        \begin{tabular*}{\textwidth}{p{0.50\textwidth}|p{0.50\textwidth}}
            \toprule
            \textbf{RegEx} & \textbf{Matching String}  \\
            \midrule
            \midrule
            \verb/Answer: [A-Z]/ & Answer: A \\
            \midrule
            \verb/Answer: \\([A-Z]\\)/ & Answer: (A) \\
            \midrule
            \verb/Answer: \\[[A-Z]\\]/ & Answer: [A] \\
            \midrule
            \verb/Answer:[A-Z]/ & Answer:A \\
            \midrule
            \verb/^[A-Z](\\.|$)/ & A. \\
            \midrule
            \verb/^\\([A-Z]\\)(\\.|$)/ & (A). \\
            \midrule
            \verb/correct answer is [A-Z]( |\\.|$)/ & ... correct answer is A. \\
            \midrule
            \verb/correct answer is \\([A-Z]\\)( |\\.|$)/ & ... correct answer is (A). \\
            \midrule
            \multirow{2}*{\texttt{correct answer is:\textbackslash n[A-Z]( |\textbackslash\textbackslash.|\$)}} & ... correct answer is: \\
                                                                     & A. \\
            \midrule
            \multirow{2}*{\texttt{correct answer is:\textbackslash n\textbackslash\textbackslash([A-Z]\textbackslash\textbackslash)( |\textbackslash\textbackslash.|\$)}}  & ... correct answer is: \\
                                                                            & (A) \\
            \midrule
            \multirow{3}*{\texttt{correct answer is:\textbackslash n\textbackslash n[A-Z]( |\textbackslash\textbackslash.|\$)}}   & ... correct answer is: \\
                                    & \\
                                    & A.\\
            \midrule
            \multirow{3}*{\texttt{correct answer is:\textbackslash n\textbackslash n\textbackslash\textbackslash([A-Z]\textbackslash\textbackslash)( |\textbackslash\textbackslash.|\$)}}   & ... correct answer is: \\
                                    & \\
                                    & (A).\\
            \midrule
            \verb/is [A-Z]( |\\.|$)/ & ... is A. \\
            \midrule
            \verb/is \\([A-Z]\\)( |\\.|$)/ & ... is (A). \\
            \midrule
            \multirow{2}*{\texttt{is:\textbackslash n[A-Z]( |\textbackslash\textbackslash.|\$)}}  & ... is: \\
                                                                                    & A. \\
            \midrule
            \multirow{2}*{\texttt{is:\textbackslash n\textbackslash\textbackslash([A-Z]\textbackslash\textbackslash)( |\textbackslash\textbackslash.|\$)}}  & ... is: \\
                                                                                    & (A). \\
            \midrule
            \multirow{3}*{\texttt{is:\textbackslash n\textbackslash n[A-Z]( |\textbackslash\textbackslash.|\$)}}  & ... is: \\
                                                                                    & \\
                                                                                    & A. \\
            \midrule
            \multirow{3}*{\texttt{is:\textbackslash n\textbackslash n\textbackslash\textbackslash([A-Z]\textbackslash\textbackslash)( |\textbackslash\textbackslash.|\$)}}  & ... is: \\
             & \\
                                                                                    & (A). \\                        
            \bottomrule
        \end{tabular*}
    \end{adjustbox}
    \caption{The 18 RegEx patterns we use to parse the LLMs outputs.}
    \label{tab:regex}
\end{table*}

\subsubsection{xFinder Models Details}
xFinder \cite{yu2024xfinderrobustpinpointanswer} is a family of models fine-tuned to extract the intended answer from generated outputs. The authors train models of varying architectures and sizes on the Key Answer Finder (KAF) dataset, which comprises question-choice samples paired with model-generated responses, specifically designed for answer extraction. In our experiments, we use their top-performing 500M and 8B models, based on Qwen1.5-0.5B and Llama-3-8B-Instruct, respectively\footnote{\url{https://huggingface.co/collections/IAAR-Shanghai/xfinder-664b7b21e94e9a93f25a8412}}.

\subsection{Prompt Details}\label{sec:appendix-prompts}

\Cref{tab:prompt-zs,tab:prompt-zs-cot,tab:prompt-zs-constr,tab:prompt-fs} present the prompts used in our experiments. When we assess performance by means of Logprobs, we also append the \textit{Assistant} tag together with the string ``\verb|Answer:|'' to the input prompt. We then look at the first-token probabilities by applying the softmax operation to the log-probability vector generated by the model for the first token. The answer choice corresponding to the token with the highest probability is selected as the predicted answer.
\begin{table}[t]
    \renewcommand{\arraystretch}{1.6}
    \centering
    \begin{tabular}{|p{0.94\linewidth}|}
        \hline
        \textbf{SYSTEM} 
        \\
        \texttt{You are an expert in question answering. Given a question and a set of choices, provide the correct answer.}
        \\
        \hline
        \textbf{USER} 
        \\
        \texttt{Question: \{question\}}
        \\[-8pt]
        \texttt{Choices: \{choices\}}
        \\
        \hline
    \end{tabular}
    \caption{Prompt for the Zero-Shot setting.}
    \label{tab:prompt-zs}
\end{table}

\begin{table}[t]
    \renewcommand{\arraystretch}{1.6}
    \centering
    \begin{tabular}{|p{0.94\linewidth}|}
        \hline
        \textbf{SYSTEM} 
        \\
        \texttt{You are an expert in question answering. Given a question and a set of choices, provide the reasoning process necessary to answer the question and then provide your answer exactly as 'Answer: [label]'.}
        \\
        \hline
        \textbf{USER} 
        \\
        \texttt{Question: \{question\}}
        \\[-8pt]
        \texttt{Choices: \{choices\}}
        \\
        \hline
    \end{tabular}
    \caption{Prompt for the Zero-Shot Chain-of-Though setting.}
    \label{tab:prompt-zs-cot}
\end{table}

\begin{table}[t]
    \renewcommand{\arraystretch}{1.6}
    \centering
    \begin{tabular}{|p{0.94\linewidth}|}
        \hline
        \textbf{SYSTEM} 
        \\
        \texttt{You are an expert in question answering. Given a question and a set of choices, provide the correct answer. Answer exactly as 'Answer: [label]'.}
        \\
        \hline
        \textbf{USER} 
        \\
        \texttt{Question: \{question\}}
        \\[-8pt]
        \texttt{Choices: \{choices\}}
        \\
        \hline
    \end{tabular}
    \caption{Prompt for the Zero-Shot Constrained setting.}
    \label{tab:prompt-zs-constr}
\end{table}

\begin{table}[t]
    \renewcommand{\arraystretch}{1.6}
    \centering
    \begin{tabular}{|p{0.94\linewidth}|}
        \hline
        \textbf{SYSTEM} 
        \\
        \texttt{You are an expert in question answering. Given a question, a set of choices, and few examples, provide the correct answer.}
        \\
        \hline
        \textbf{USER} 
        \\
        \texttt{Question: \{question\}}
        \\[-8pt]
        \texttt{Choices: \{choices\}}
        \\
        \hline
    \end{tabular}
    \caption{Prompt for the Few-Shot setting.}
    \label{tab:prompt-fs}
\end{table}


\subsection{Annotation Guidelines and Invalid Answer Examples}\label{sec:appendix-ann}

\begin{table*}[t]
    \begin{adjustbox}{width=\textwidth}
    \footnotesize
    \centering
    \begin{tabular*}{\textwidth}{p{0.15\textwidth}|p{0.25\textwidth}|p{0.55\textwidth}}
        \toprule
        \textbf{Error type} & \textbf{Description} & \textbf{Example} \\ 
        \midrule
        \parbox[t]{0.15\textwidth}{Conflicting answer}
        &
        \parbox[t]{0.25\textwidth}{The model output includes conflicting answers, such as inconsistent reasoning or multiple answers.}
        &
        \parbox[t]{0.53\textwidth}{
        \textbf{Question}: When light undergoes interference, it can sometimes
        \textbf{Choices}: A. build up to more than the sum of amplitudes B. cancel completely C. Both of these D. Neither of these\\
        \textbf{Model Output}: When light waves overlap, \colorbox{nicegreen}{they can build} \colorbox{nicegreen}{up to more than the sum of amplitudes (addition) or cancel each} \colorbox{nicegreen}{other out (destruction)}. Therefore, the correct answer is \colorbox{nicered}{(A) build up to more than the sum of amplitudes}.
        }\\
        \midrule
        \parbox[t]{0.15\textwidth}{Inconsistency in label binding}        
        & 
        \parbox[t]{0.25\textwidth}{The model fails to bind the choice's label to its corresponding text.}        
        &
        \parbox[t]{0.55\textwidth}{
        \textbf{Question}: What is the capital of France?\\  
        \textbf{Choices}: A) Berlin, B) Madrid, \colorbox{nicegreen}{C) Paris}, D) Rome\\ 
        \textbf{Model Output}: The capital of France is \colorbox{nicered}{A. Paris}.
        } \\
        \midrule
        \parbox[t]{0.15\textwidth}{Refused to answer}
        &
        \parbox[t]{0.25\textwidth}{The model explicitly states that it cannot answer the question.}
        &
        \parbox[t]{0.53\textwidth}{
        \textbf{Question}: How much energy do Americans waste each year by throwing out the unused ice in their drinks?\\
        \textbf{Choices}: 
        A. $6\times10^8$ J
        B. $6\times10^{11}$ J
        C. $6\times10^{14}$ J
        D. $6\times10^{17}$ J\\
        \textbf{Model Output}: \colorbox{nicered}{I couldn't find any information} about the amount of energy wasted by throwing out unused ice in drinks. This seems to be an unknown or fictional fact.
        }\\
        \midrule
        \parbox[t]{0.15\textwidth}{Irrelevant response}
        &
        \parbox[t]{0.25\textwidth}{The model fails to engage with the question or generates an output which is not relevant to the question.}
        &
        \parbox[t]{0.55\textwidth}{
        \textbf{Question}: Sofia and Tess will each randomly choose one of the 10 integers from 1 to 10. What is the probability that neither integer chosen will be the square of the other?\\  
        \textbf{Choices}: A. 0.64 $\quad$ B. 0.72 $\quad$ C. 0.81 $\quad$ D. 0.95\\
        \textbf{Model Output}: To find the probability that neither integer chosen will be the square of the other, we need to consider the cases where this condition is not met and subtract it from 1.
        [...]
        The probability that neither integer chosen will be the square of the other is 1 - (9/100) = 91/100 = 0.91.
        \colorbox{nicered}{However, none of the given choices match this answer.}
        }\\
        \midrule
        \parbox[t]{0.15\textwidth}{Reach token limit}
        &
        \parbox[t]{0.25\textwidth}{The model output is truncated as it reached the maximum token limit.}
        &
        \parbox[t]{0.53\textwidth}{
        \textbf{Question}: The language {ww | w in (0 + 1)*} is\\
        \textbf{Choices}: A. not accepted by any Turing machine B. accepted by some Turing machine, but by no pushdown automaton C. accepted by some pushdown automaton, but not context-free D. context-free, but not regular\\
        \textbf{Model Output}: Great! Let's break down the question and the choices: [...] The language {ww | w in (0 + 1)*} is context-sensitive, but not context \colorbox{nicered}{\texttt{[EOS]}}
        }\\
        \bottomrule
    \end{tabular*}
    \end{adjustbox}
    \caption{Identified error categories in model outputs with corresponding examples. Errors are highlighted in red, while input cues for correct answers (if present) are shown in green. Examples are drawn from the MMLU-Redux dataset post-annotation. A similar table was provided to annotators before the annotation process.}
    \label{tab:invalid-answers}
\end{table*}

In this section, we outline the guidelines followed by our annotators for the MMLU-Redux dataset and present statistics on annotated invalid answer types. 
The goal of the annotation process is to manually identify the model's intended answer from its generated output.
The annotation process was conducted by four expert Ph.D. students annotators, all possessing at least a C1 level of English proficiency.

For each sample, human annotators carefully reviewed the question, answer choices, and model output before selecting the intended answer (i.e., ``A'', ``B'', ``C'', ``D'', or ``[No valid answer]''). 
Annotators did not have access to the ground-truth answer or information about the model that generated the response. They were instructed to accept both explicit answers (e.g., ``The correct answer is B'') and implicit ones, provided the reasoning was coherent.

\subsubsection{Annotation Procedure}

Annotators followed a structured process:

\paragraph{Step 1: Read the Question and Answer Choices}  
Understand the question's context and review all answer choices (A, B, C, D) to ensure clarity.

\paragraph{Step 2: Read the Model Output}  
Analyze the entire response, considering explicit answers, reasoning leading to an answer, and any conflicting statements.

\paragraph{Step 3: Extract the Intended Answer}  
Identify the direct answer or infer it from the model's reasoning if unstated. Cases where the response is ambiguous, irrelevant, or contradictory are labeled as ``[No valid answer].''

As discussed in \Cref{sec:results}, we define five types of outputs that should be marked as ``[No valid answer]'' and include them in the annotation guidelines with examples:

\paragraph{Conflicting Answer (C.A.):}The model output exhibits conflicting reasoning. For instance, it may suggest that multiple answers are correct or present a reasoning pattern that contradicts the explicitly stated answer.
\paragraph{Inconsistency in Label Binding (I.L.B.):} The model associates a choice's text with the wrong label, e.g., generating ``A. 3'' when the original choices were ``A. 4'' and ``C. 3''.
\paragraph{Refused to Answer (R.A.):} The model explicitly declines to answer due to insufficient knowledge or safety concerns.
\paragraph{Irrelevant Response (I.R.):} The model references choices not provided (e.g., ``The answer is E'' when E is not an option), or states that there is no valid choice among the ones provided.
\paragraph{Reached Token Limit (R.T.L.):} The model fails to generate a complete response within the available 512 tokens, making it impossible to infer the intended answer.

\Cref{tab:invalid-answers} provides examples of these invalid answer patterns. \Cref{tab:invalid-answers-notes} presents the number of instances tagged as invalid answers across different prompt settings for the $1,000$ manually annotated samples. We observe that the frequency of ``[No valid answer]'' varies with the prompt setting. Notably, under ZS and ZS-CoT settings, outputs are more prone to invalid answers. The most common invalid answer type is ``\textit{conflicting answer}'' (C.A.).

\begin{table}[t]
    \centering
    \begin{adjustbox}{max width=\columnwidth}
    \centering
        \begin{tabular}{c|c|c|c|c|c|c}
            \toprule
            \textbf{Setting} & \textbf{C.A.} & \textbf{I.L.B} & \textbf{R.A.} & \textbf{I.R.} & \textbf{R.T.L} & \textbf{\# Samples} \\
            \midrule
            \textbf{ZS}       & 5  & 2  & 1 & 3  & 5  & 250 \\
            \textbf{ZS-CoT}   & 11 & 0  & 1 & 7  & 10 & 250 \\
            \textbf{ZS-Const} & 3  & 0  & 1 & 1  & 0  & 250 \\
            \textbf{FS}       & 4  & 0  & 3 & 1  & 0  & 250 \\
            \midrule
            \textbf{All}      & 23 & 2  & 6 & 12 & 15 & 1000 \\
            \bottomrule
        \end{tabular}
    \end{adjustbox}
    \caption{Number of instances annotated as ``[No valid answer]'' for each prompt setting.}
    \label{tab:invalid-answers-notes}
\end{table}

\subsection{Results on ARC}\label{sec:appendix-arc}
\begin{figure}[t]
    \centering
    \includegraphics[width=0.95\columnwidth]{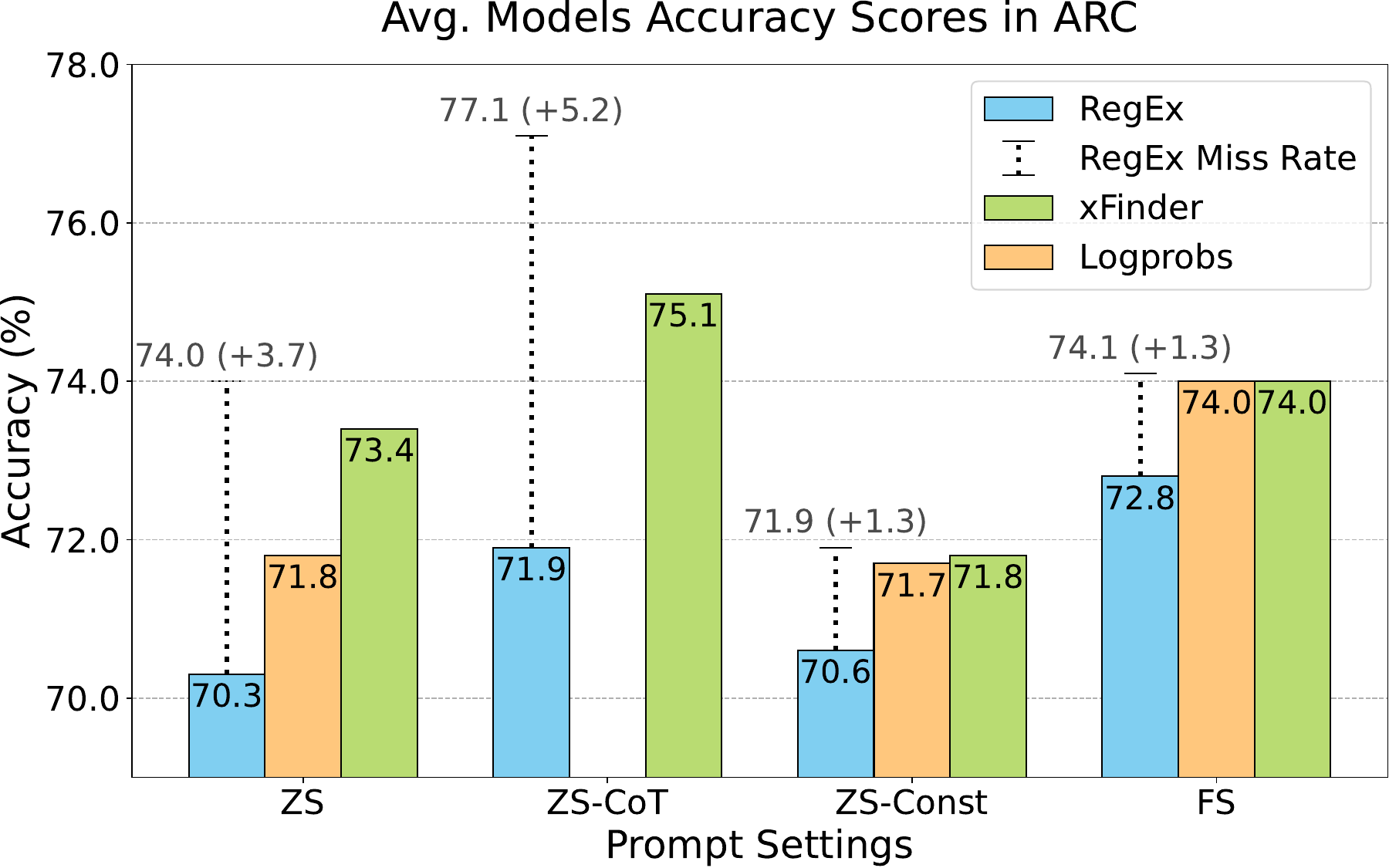}
    \caption{Average accuracy scores across eight LLMs and four prompt settings when evaluated on the ARC dataset. Dotted lines indicate the RegEx miss rate.}
    \label{img:arc}
\end{figure}
\begin{figure}[t]
    \centering
    \includegraphics[width=\columnwidth]{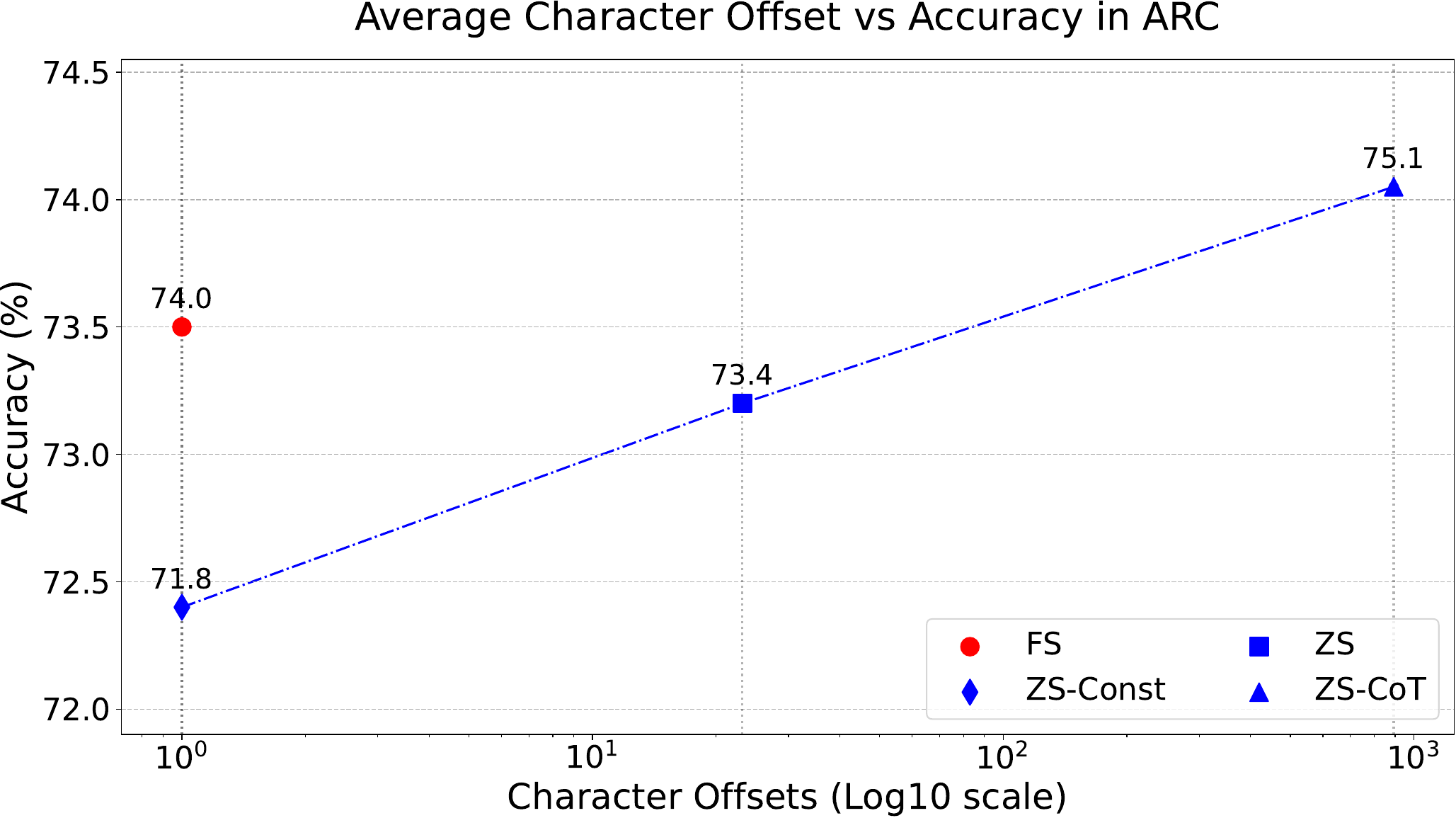}    
    \caption{The plot shows the average answer offset ($\log_{10}$ scale) and accuracy (\%) across the four different prompt setting using the xFinder evaluation strategy.}
    \label{img:offset-arc}
\end{figure}

In this section, we present the results for the ARC dataset across different evaluation strategies and prompt settings.

From \Cref{img:arc}, we can observe that, since ARC consists of grade-school science questions, model performance closely aligns with the results obtained for the \textsc{Stem} category in MMLU-Redux (\Cref{img:stem}). Despite high miss rates, models evaluated with the RegEx strategy in the ZS and ZS-CoT settings perform comparably or even better than in the ZS-Const setting, underscoring the benefits of allowing free-form generation.
This trend is further supported by the xFinder-based evaluation, where both ZS and ZS-CoT outperform ZS-Const, and ZS-CoT even surpasses FS. To reinforce these findings, \Cref{img:offset-arc} reports average answer offsets across prompt settings. The higher offset values in ZS-CoT correlate strongly with final model performance, except in FS, which we believe compensates for the lack of explicit reasoning by leveraging the benefits of learning from in-context examples.

\subsection{Individual LLM Results}\label{sec:appendix-each-llm}
In order to have a more comprehensive analysis, we present the accuracy results separately for each of the LLMs under investigation (\Cref{sec:models}). The results are organized into three distinct tables: \Cref{tab:mmlu-single-results} for MMLU-Redux, \Cref{tab:obqa-single-results} for OBQA, and \Cref{tab:arc-single-results} for ARC.

The reported results validate the consistency of our findings across all tested models, further strengthening the conclusions highlighted in \Cref{sec:results}.

\begin{table*}[t]
    \centering
    \begin{adjustbox}{max width=\textwidth}
    \centering
        \begin{tabular}{l|ccc|ccc|ccc|ccc}
            \toprule
             & \multicolumn{3}{c}{\textbf{ZS}} & \multicolumn{3}{c}{\textbf{ZS-CoT}} & \multicolumn{3}{c}{\textbf{ZS-Const}} & \multicolumn{3}{c}{\textbf{FS}} \\
            \cmidrule{2-13}
            \textbf{Model} & \textbf{RegEx} & \textbf{Logprobs} & \textbf{xFinder} & \textbf{RegEx} & \textbf{Logprobs} & \textbf{xFinder} & \textbf{RegEx} & \textbf{Logprobs} & \textbf{xFinder} & \textbf{RegEx} & \textbf{Logprobs} & \textbf{xFinder}\\
            \midrule
            \textbf{Qwen2.5-7B-Instruct} & 71.4 & 72.1 & 73.5 & 75.2 & --- & 75.8 & 72.1 & 72.1 & 72.1 & 74.0 & 74.0 & 74.1\\
            \midrule
            \textbf{Llama-3.1-8B-Instruct} & 67.3 & 67.0 & 69.1 & 70.6 & --- & 71.6 & 66.8 & 66.8 & 66.8 & 68.8 & 69.0 & 69.8\\
            \midrule
            \textbf{Mistral-v0.3-7B-Instruct} & 55.2 & 57.4 & 57.1 & 53.0 & --- & 58.0 & 57.4 & 58.1 & 58.1 & 59.5 & 59.8 & 59.7\\
            \midrule
            \textbf{Llama-2-7b-chat-hf} & 44.0 & 44.4 & 46.1 & 37.7 & --- & 43.7 & 43.1 & 43.4 & 43.3 & 48.5 & 49.5 & 48.7\\
            \midrule
            \textbf{Phi-4-mini-instruct} & 68.0 & 66.4 & 68.6 & 70.2 & --- & 70.8 & 65.4 & 65.7 & 65.8 & 68.6 & 68.6 & 68.6\\
            \midrule
            \textbf{Phi-3.5-mini-instruct} & 67.2 & 67.9 & 70.2 & 70.1 & --- & 70.9 & 67.8 & 67.4 & 68.4 & 67.9 & 67.7 & 69.4\\
            \midrule
            \textbf{Llama-3.2-1B-Instruct} & 44.6 & 45.6 & 47.5 & 38.8 & --- & 44.9 & 44.2 & 44.0 & 44.3 & 47.7 & 48.1 & 47.9\\
            \midrule
            \textbf{Smol-2-1.7B-Instruct} & 44.7 & 45.1 & 45.5 & 45.2 & ---& 46.7 & 44.5 & 44.4 & 44.7 & 474 & 47.3 & 47.4\\
            \bottomrule
        \end{tabular}
    \end{adjustbox}
    \caption{Accuracy results on MMLU-Redux showing the individual performance of each model across four prompt settings and three evaluation strategies.}
    \label{tab:mmlu-single-results}
\end{table*}

\begin{table*}[t]
    \centering
    \begin{adjustbox}{max width=\textwidth}
    \centering
        \begin{tabular}{l|ccc|ccc|ccc|ccc}
            \toprule
             & \multicolumn{3}{c}{\textbf{ZS}} & \multicolumn{3}{c}{\textbf{ZS-CoT}} & \multicolumn{3}{c}{\textbf{ZS-Const}} & \multicolumn{3}{c}{\textbf{FS}} \\
            \cmidrule{2-13}
            \textbf{Model} & \textbf{RegEx} & \textbf{Logprobs} & \textbf{xFinder} &  \textbf{RegEx} & \textbf{Logprobs} & \textbf{xFinder} &  \textbf{RegEx} & \textbf{Logprobs} & \textbf{xFinder} &  \textbf{RegEx} & \textbf{Logprobs} & \textbf{xFinder} \\
            \midrule
            \textbf{Qwen2.5-7B-Instruct} & 77.0 & 87 & 85.2 & 88.8 & --- & 89.0 & 87.2 & 87.2 & 87.4 & 89.4 & 90.2 & 89.4\\
            \midrule
            \textbf{Llama-3.1-8B-Instruct} & 81.4 & 82.6 & 81.4 & 85.4 & --- & 85.4 & 82.6 & 83.0 & 82.6 & 81.0 & 81.0 & 81.0\\
            \midrule
            \textbf{Mistral-v0.3-7B-Instruct} & 66.6 & 66.6 & 70.0 & 67.6 & --- & 71.6 & 71.0 & 71.6 & 70.8 & 74.2 & 74.4 & 74.0\\
            \midrule
            \textbf{Llama-2-7b-chat-hf} & 57.2 & 52.4 & 57.6 & 53.2 & --- & 54.6 & 52.2 & 52.0 & 52.6 & 63.2 & 64.2 & 63.2\\
            \midrule
            \textbf{Phi-4-mini-instruct} & 80.2 & 81.4 & 80.4 & 81.8 & --- & 82.4 & 82.6 & 82.0 & 82.8 & 83.8 & 83.8 & 83.8\\
            \midrule
            \textbf{Phi-3.5-mini-instruct} & 80.0 & 83.4 & 83.0 & 85.2 & --- & 85.6 & 84.0 & 83.8 & 84.0 & 86.0 & 85.2 & 86.0\\
            \midrule
            \textbf{Llama-3.2-1B-Instruct} & 55.8 & 51.0 & 56.6 & 51.2 & --- & 55.4 & 54.8 & 55.0 & 54.8 & 60.0 & 60.0 & 60.0\\
            \midrule
            \textbf{Smol-2-1.7B-Instruct} & 54.0 & 54.4 & 54.8 & 55.0 & --- & 56.4 & 54.0 & 54.8 & 54.2 & 59.0 & 59.4 & 59.0 \\
            \bottomrule
        \end{tabular}
    \end{adjustbox}
    \caption{Accuracy results on OBQA showing the individual performance of each model across four prompt settings and three evaluation strategies.}
    \label{tab:obqa-single-results}
\end{table*}

\begin{table*}[!th]
    \centering
    \begin{adjustbox}{max width=\textwidth}
    \centering
        \begin{tabular}{l|ccc|ccc|ccc|ccc}
            \toprule
             & \multicolumn{3}{c}{\textbf{ZS}} & \multicolumn{3}{c}{\textbf{ZS-CoT}} & \multicolumn{3}{c}{\textbf{ZS-Const}} & \multicolumn{3}{c}{\textbf{FS}} \\
            \cmidrule{2-13}
            \textbf{Model} & \textbf{RegEx} & \textbf{Logprobs} & \textbf{xFinder} &  \textbf{RegEx} & \textbf{Logprobs} & \textbf{xFinder} &  \textbf{RegEx} & \textbf{Logprobs} & \textbf{xFinder} &  \textbf{RegEx} & \textbf{Logprobs} & \textbf{xFinder} \\
            \midrule
            \textbf{Qwen2.5-7B-Instruct} & 85.4 & 89.3 & 89.5 & 89.7 & --- & 91.7 & 87.7 & 89.5 & 89.5 & 87.8 & 89.7 & 89.6\\
            \midrule
            \textbf{Llama-3.1-8B-Instruct} & 81.8 & 82.4 & 84.2 & 85.8 & --- & 87.7 & 80.6 & 82.3 & 82.3 & 80.9 & 82.4 & 83.1\\
            \midrule
            \textbf{Mistral-v0.3-7B-Instruct} & 70.6 & 73.9 & 74.1 & 72.6 & --- & 75.1 & 72.9 & 73.0 & 73.5 & 75.9 & 76.9 & 76.1\\
            \midrule
            \textbf{Llama-2-7b-chat-hf} & 54.5 & 52.2 & 57.1 & 49.5 & --- & 55.3 & 50.2 & 52.3 & 52.6 & 59.5 & 61.2 & 61.1\\
            \midrule
            \textbf{Phi-4-mini-instruct} & 82.4 & 82.8 & 82.8 & 86.8 & --- & 87.0 & 82.8 & 83.0 & 82.9 & 83.0 & 83.1 & 83.0 \\
            \midrule
            \textbf{Phi-3.5-mini-instruct} & 81.8 & 86.5 & 87.3 & 86.8 & --- & 88.9 & 84.3 & 85.4 & 86.6 & 84.3 & 86.0 & 86.6\\
            \midrule
            \textbf{Llama-3.2-1B-Instruct} & 50.1 & 51.7 & 54.6 & 47.0 & --- & 54.8 & 51.2 & 52.3 & 52.3 & 54.1 & 55.7 & 55.7\\
            \midrule
            \textbf{Smol-2-1.7B-Instruct} & 56.0 & 55.2 & 57.2 & 57.0 & --- & 59.6 & 54.4 & 55.3 & 54.6 & 56.9 & 57.0 & 56.8 \\
            \bottomrule
        \end{tabular}
    \end{adjustbox}
    \caption{Accuracy results on ARC showing the individual performance of each model across four prompt settings and three evaluation strategies.}
    \label{tab:arc-single-results}
\end{table*}

\subsection{Additional Results on MMLU-Redux}\label{sec:appendix-sub}

\begin{figure}[t]
    \centering
    \begin{subfigure}{\columnwidth}
        \centering
        \includegraphics[width=\columnwidth]{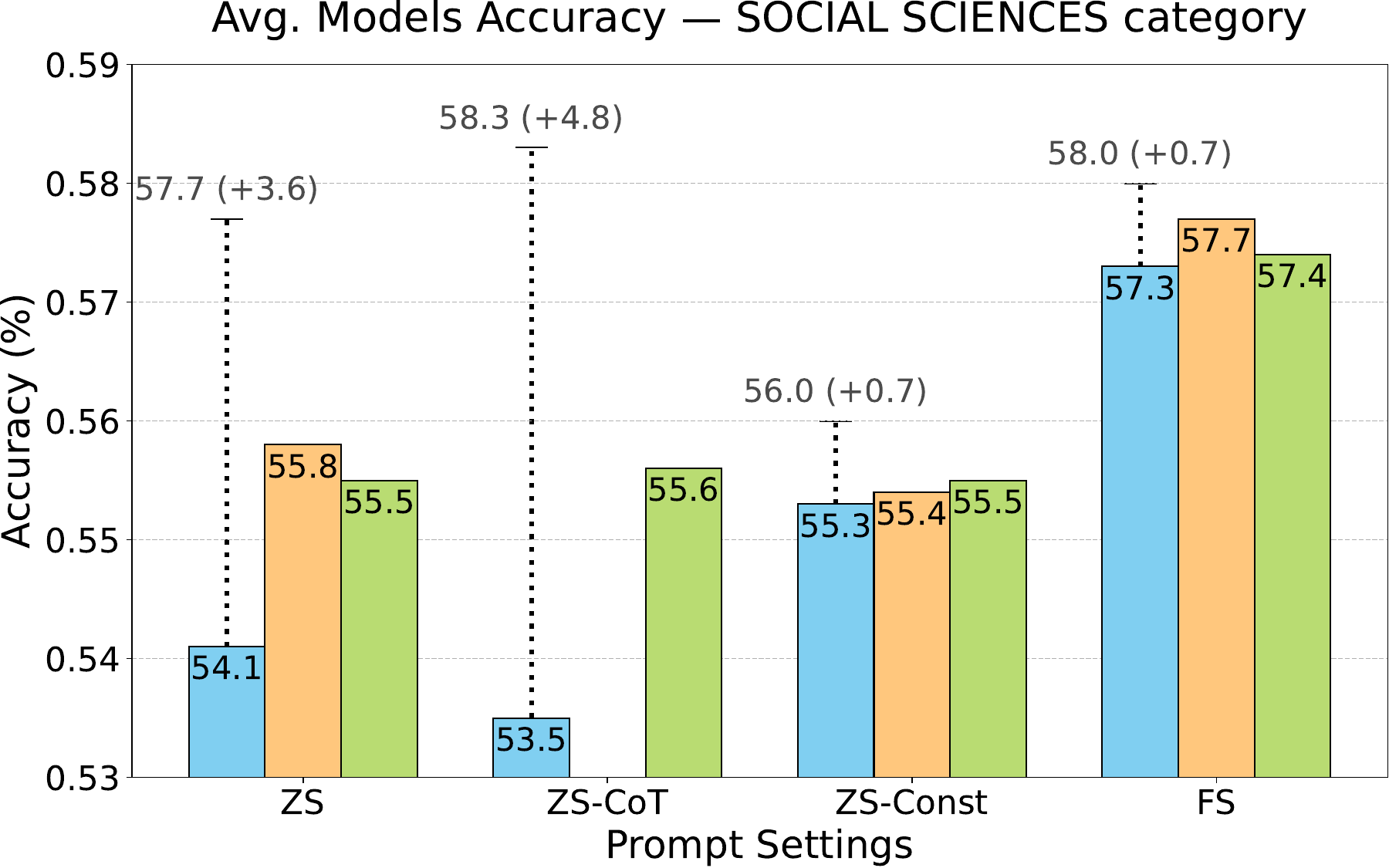}
        \caption{\textsc{Social Sciences} category}
        \label{img:social_sciences}
    \end{subfigure}
    
    \vspace{1em}
    
    \begin{subfigure}{\columnwidth}
        \centering
        \includegraphics[width=\columnwidth]{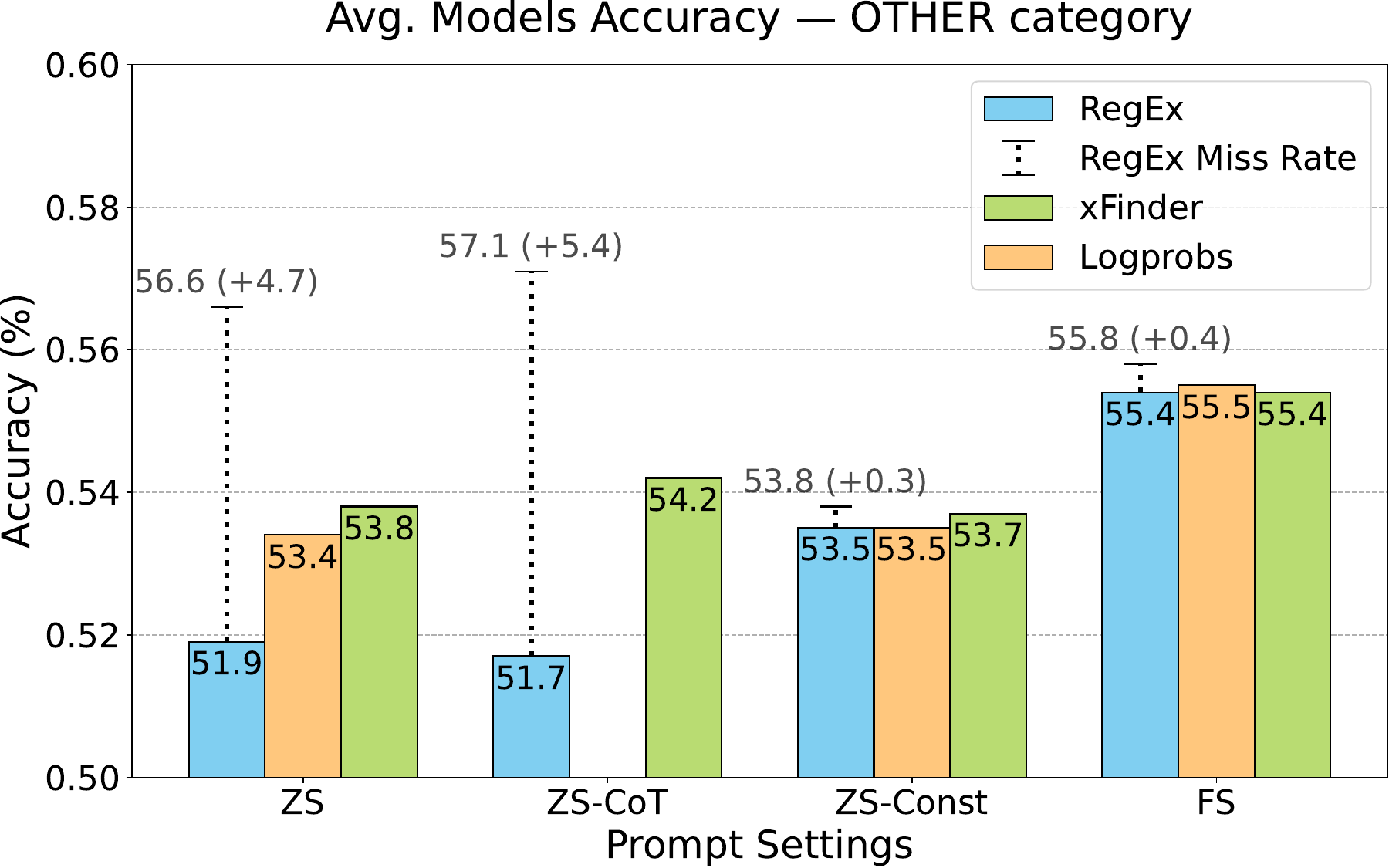}
        \caption{\textsc{Other} category}
        \label{img:other}
    \end{subfigure}
    
    \caption{Average accuracy scores across eight LLMs and four prompt settings when evaluated on the \textsc{Social Sciences} (\Cref{img:social_sciences}) and \textsc{Other} (\Cref{img:other}) categories of MMLU-Redux. Dotted lines indicate the RegEx miss rate.}
    \label{fig:social_other}
\end{figure}

In this section, we present additional results for the last two main categories of MMLU-Redux: \textsc{Social Sciences} and \textsc{Other} (\Cref{img:social_sciences,img:other}).  

As shown in the figures, model performance on these categories follow a trend similar to that of the \textsc{Humanities} category (\Cref{img:humanities}). This is expected, as they include subcategories such as \textsc{Global Facts}, \textsc{Business Ethics}, \textsc{High School Geography}, and \textsc{High School Psychology}, among others, which are less aligned with \textsc{Stem} fields like \textsc{Abstract Algebra} and \textsc{College Mathematics}.  

\subsection{MMLU-Adversarial Examples and Prompts}\label{sec:appendix-mmlu-adv}
\Cref{tab:inconsistent-example} and \Cref{tab:multiple-answers-example} illustrate examples of the inconsistent reasoning and multiple answers error patterns, respectively. In each table, the first row shows a real model output that was manually annotated as exhibiting the corresponding error. These examples serve as in-context demonstrations to guide Gemini-1.5-Flash during generation. The second row differs by error type: for inconsistent reasoning, it includes an original model output and an adversarially generated output from Gemini that reproduces the same reasoning while selecting a different answer. For multiple answers, the second row contains only the adversarial output, which presents a rationale that supports more than one possible answer. \Cref{tab:system-prompt-inconsistent} and \Cref{tab:system-prompt-multiple} show the system prompts used to generate adversarial examples for the inconsistent reasoning and multiple answers error patterns, respectively.
\subsection{Answer Extractors Adversarial Prompts}\label{sec:appendix-adv}
\Cref{tab:prompt-xfinder} shows the adversarial prompts used to stimulate xFinder to perform the MCQA task. Specifically, Prompt A and Prompt B list all possible answers and state that one is correct without specifying which one, while Prompt C simply lists all available choices in natural order.

\begin{table}[t]
    \renewcommand{\arraystretch}{1.6}
    \centering
    \begin{tabular}{|p{0.94\linewidth}|}
        \hline
        \textbf{Prompt A} 
        \\
        \texttt{Given the Answer Choices: \{choices\}. There is a correct option, and it is quite straightforward to identify in}
        \\
        \hline
        \textbf{Prompt B} 
        \\
        \texttt{Given the Answer Choices: \{choices\}. There is a right choice, and it is fairly easy to be recognized in}
        \\
        \hline
        \textbf{Prompt C} 
        \\
        \texttt{The correct option is: \{choices\}}
        \\
        \hline
    \end{tabular}
    \caption{Prompt configuration for the xFinder biasing test. The \texttt{\{choices\}} string serves as a placeholder for the list of answer choices, which is instantiated for each sample.}
    \label{tab:prompt-xfinder}
\end{table}
\begin{table}[t]
    \renewcommand{\arraystretch}{1.6}
    \centering
    \begin{tabular}{|p{0.94\linewidth}|}
        \hline
        \textbf{System Prompt} 
        \\
        \texttt{Given an input question, answer choices, and an output generated by a model, substitute ONLY the model answer label and text with another one among the ones available. DO NOT change anything else in the output.}
        \\
        \hline
    \end{tabular}
    \caption{System prompt used with Gemini-1.5-Flash, to create the Inconsistent Reasoning subset of MMLU-Adversarial}
    \label{tab:system-prompt-inconsistent}
\end{table}

\begin{table}[t]
    \renewcommand{\arraystretch}{1.6}
    \centering
    \begin{tabular}{|p{0.94\linewidth}|}
        \hline
        \textbf{System Prompt} 
        \\
        \texttt{You are an expert at the game of writing convincing pseudo-explanation. Given an input question and an answer, write the explanation for why that answer is correct with respect to the question. If the answer seems not correct to you, do not acknowledge it. End your explanation with 'Answer: [label].'}
        \\
        \hline
    \end{tabular}
    \caption{System prompt used with Gemini-1.5-Flash to create the Multiple Answers subset of MMLU-Adversarial}
    \label{tab:system-prompt-multiple}
\end{table}

\begin{table*}[t]
    \begin{adjustbox}{width=\textwidth}
    \centering
    \footnotesize
    \begin{tabular*}{\textwidth}{p{0.40\textwidth}|p{0.55\textwidth}}
        \toprule
        \parbox[t]{0.40\textwidth}{\textbf{Input sample}\\ Question: On which continent are most of the Venus figurines found?\\Choices:\\A. Europe\\B. Africa\\C. Asia\\D. Australia}        
        &
        \parbox[t]{0.55\textwidth}{
        \textbf{Original Model Output}\\
        \texttt{[...] \colorbox{niceblue}{The majority of Venus figurines have been found on} \colorbox{niceblue}{the continent of Europe}.
        In fact, over 70\% of all known Venus figurines are from Europe, with the majority coming from the Upper Paleolithic period (around 40,000-10,000 BCE). [...]\\ 
        Therefore, the answer to the question is: \colorbox{nicered}{Answer: B. Africa}}
        }\\
        \midrule
        \parbox[t]{0.40\textwidth}{
        \textbf{Input sample}\\
        Question: How much toothpaste is recommended for cleaning teeth?\\
        Choices:\\
        A. A smear.\\
        B. A pea-sized amount.\\
        C. The length of the toothbrush.\\
        D. Half an inch.\\
        }
        &
        \parbox[t]{0.55\textwidth}{
        \textbf{Original Model Output}\\
        \texttt{The correct answer is \colorbox{niceblue}{B. A pea-sized amount.}\\
        According to the American Dental Association (ADA) and most dental professionals, \colorbox{niceblue}{a pea-sized amount of} \colorbox{niceblue}{toothpaste is recommended for cleaning teeth}, especially for children.}
        \\[0.5em]
        \textbf{Adversarial Model Output (\textsc{Gemini-1.5-Flash})}\\
        \texttt{The correct answer is \colorbox{nicered}{A. A smear}.\\ According to the American Dental Association (ADA) and most dental professionals, \colorbox{niceblue}{a pea-sized amount of} \colorbox{niceblue}{toothpaste is recommended for cleaning teeth}, especially for children.}
        }\\

        \bottomrule
    \end{tabular*}
    \end{adjustbox}
    \caption{Examples of outputs showing inconsistent reasoning (blue highlights the reasoning path, while red highlights the answer given as correct).
    The first row shows one of the annotated samples, while the second row depicts a coherent sample generated by one of the models used in our study, along with the relative adversarial sample generated by Gemini-1.5-Flash.}
    \label{tab:inconsistent-example}
\end{table*}

\begin{table*}[t]
    \begin{adjustbox}{width=\textwidth}
    \centering
    \footnotesize
    \begin{tabular*}{\textwidth}{p{0.40\textwidth}|p{0.55\textwidth}}
        \toprule
        \parbox[t]{0.40\textwidth}{
        \textbf{Input sample}\\
        Question: If $f (x) = x^3 + 2x^2 - 9x - 18$, which of the following statements is true?\\
        Choices:\\ 
        A. $f (x) = 0$ has three real solutions.\\ 
        B. $f (x) \geq -18$ for all $x \geq 0$.\\ 
        C. $f (x) \leq -18$ for all $x \leq 0$.\\
        D. The function $f (x)$ is decreasing for $x \leq -3$.}
        &
        \parbox[t]{0.55\textwidth}{
        \textbf{Original Model Output}\\
        \texttt{To determine the correct answer, we need to analyze the given function.\\
        A. f(x) = 0 has three real solutions: [...]  
        So, \colorbox{nicered}{the correct answer is A}.\\  
        B. f(x) >= -18 for all x >= 0:  We can rewrite the function as [...] 
        Since 1782 is greater than -18, \colorbox{nicered}{the correct answer is B}.\\  
        C. f(x) <= -18 for all x <= 0:  We can rewrite the function as [...] 
        Since -64 is less than -18, the \colorbox{nicered}{correct answer is C}.\\  
        D. The function f(x) is decreasing for x <= -3:  We can determine [...] 
        \colorbox{nicered}{the correct answer is D}.}}\\
        \midrule
        \parbox[t]{0.40\textwidth}{
        \textbf{Input sample}\\
        What number makes the equation 35 / ? = 7 true?\\
        A. 5\\
        B. 6\\
        C. 7\\
        D. 8}
        &
        \parbox[t]{0.55\textwidth}{
        \textbf{Adversarial Model Output (\textsc{Gemini-1.5-Flash})}\\
        \texttt{\colorbox{nicered}{C. 7 is the correct answer} because 35 divided by a divisor that is a multiple of 7 will yield a quotient of 7. 
        [...] \colorbox{nicered}{Answer: C}.
        \\ 
        \colorbox{nicered}{A. 5}.
        What makes it valid is that \colorbox{nicered}{35 divided by 5 equals 7}, which is a fundamental property of arithmetic division. 
        [...] and it's what makes 5 the correct answer.}
        }\\
        \bottomrule
    \end{tabular*}
    \end{adjustbox}
    \caption{Examples of outputs showing multiple answers (red highlights multiple answers given as correct).
    The first sample is one of the annotated samples, labeled as ``A'' by xFinder, while the second is an adversarial sample generated by Gemini-1.5-Flash.}
    \label{tab:multiple-answers-example}
\end{table*}

\end{document}